\documentclass{ecai} 

\usepackage{latexsym}
\usepackage{amssymb}
\usepackage{amsmath}
\usepackage{amsthm}
\usepackage{booktabs}
\usepackage{enumitem}
\usepackage{graphicx}
\usepackage{xcolor}

\usepackage{microtype}
\usepackage{subcaption}
\usepackage{multirow}
\usepackage{rotating}
\usepackage{pdflscape}
\usepackage{hyperref}

\def\our{PPCEF}

\begin{document}
\begin{frontmatter}

%%% Use this command to specify your submission number.
%%% In doubleblind mode, it will be printed on the first page.

\paperid{1214} 

\title{Probabilistically Plausible Counterfactual Explanations with Normalizing Flows}

%%% Use this combinations of commands to specify all authors of your 
%%% paper. Use \fnms{} and \snm{} to indicate everyone's first names 
%%% and surname. This will help the publisher with indexing the 
%%% proceedings. Please use a reasonable approximation in case your 
%%% name does not neatly split into "first names" and "surname".
%%% Specifying your ORCID digital identifier is optional. 
%%% Use the \thanks{} command to indicate one or more corresponding 
%%% authors and their email address(es). If so desired, you can specify
%%% author contributions using the \footnote{} command.

\author[A]{\fnms{Patryk}~\snm{Wielopolski}\thanks{Corresponding Author. Email: patryk.wielopolski@pwr.edu.pl.}}
\author[A]{\fnms{Oleksii}~\snm{Furman}}
\author[B]{\fnms{Jerzy}~\snm{Stefanowski}}
\author[A,C]{\fnms{Maciej}~\snm{Zięba}}

\address[A]{Wrocław University of Science and Technology}
\address[B]{Poznań University of Technology}
\address[C]{Tooploox Sp. z o.o.}

\begin{abstract}
We present \our{}, a novel method for generating probabilistically plausible counterfactual explanations (CFs). \our{} advances beyond existing methods by combining a probabilistic formulation that leverages the data distribution with the optimization of plausibility within a unified framework. Compared to reference approaches, our method enforces plausibility by directly optimizing the explicit density function without assuming a particular family of parametrized distributions. This ensures CFs are not only valid (i.e., achieve class change) but also align with the underlying data's probability density. For that purpose, our approach leverages normalizing flows as powerful density estimators to capture the complex high-dimensional data distribution. Furthermore, we introduce a novel loss function that balances the trade-off between achieving class change and maintaining closeness to the original instance while also incorporating a probabilistic plausibility term. \our{}'s unconstrained formulation allows for an efficient gradient-based optimization with batch processing, leading to orders of magnitude faster computation compared to prior methods. Moreover, the unconstrained formulation of \our{} allows for the seamless integration of future constraints tailored to specific counterfactual properties. Finally, extensive evaluations demonstrate \our{}'s superiority in generating high-quality, probabilistically plausible counterfactual explanations in high-dimensional tabular settings. % This makes \our{} a powerful tool for not only interpreting complex machine learning models but also for improving fairness, accountability, and trust in AI systems.
\end{abstract}
\end{frontmatter}

%%%%%%%%%%%%%%%%%%%%%%%%%%%%%%%%%%%%%%%%%%%%%%%%%%%%%%%%%%%%%%%%%%%%%%%%

\section{Introduction}

Counterfactual explanations (briefly \textit{counterfactuals}, and abbreviated as CF) are one particular type of such explanations of black box model predictions that provide information about how feature values of an example should be changed to obtain a more desired prediction of the model (i.e., to change its target decision)~\cite{Verma2020}. On the one hand, by interacting with the model using counterfactuals, the user can better understand how the system works by exploring "what would have happened if..." scenarios. On the other hand, a good counterfactual provides a practical recommendation to the user about what changes are needed in order to achieve the desired outcome. 

There are many practical applications for counterfactual explanations, including loan or insurance decisions~\cite{WachterMR17}, recruitment processes~\cite{pearl2016causal}, the discovery of chemical compounds~\cite{counterfactuals_chemistry}, medical diagnosis~\cite{counterfactuals_medical}, and many others, see, e.g., the recent survey~\cite{Guidotti22}.

\begin{figure}[t]
\begin{center}
\centerline{
    \includegraphics[width=\columnwidth]{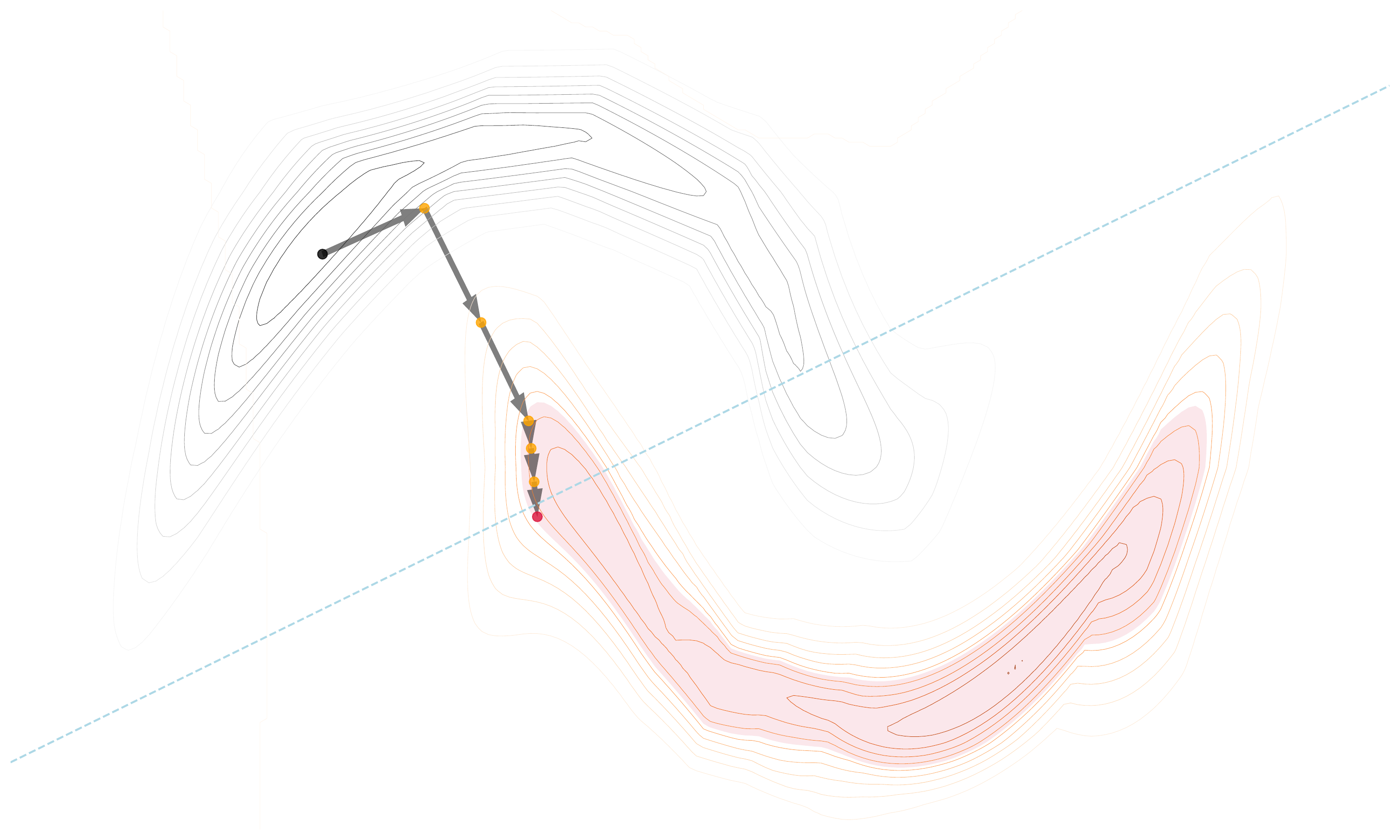}
}
\caption{Probabilistically Plausibile Counterfactual Explanation Estimation Process on the Moons Dataset. We show an evolution of an instance from the initial instance (black dot) to the final counterfactual (red dot) against the linear classifier's decision boundary (blue line) and density threshold contours, highlighting the method's trajectory towards achieving target classification and probabilistic plausibility condition.}
\label{fig:moons}
\end{center}
\vskip 0.3in
\end{figure}

More formally, a counterfactual explanation is an alternative input instance, denoted as $\mathbf{x}'$, which is minimally modified from the description of the original instance $\mathbf{x}_0$, such that the output of the classifier $h$ changes from the original decision $y = h(\mathbf{x}_0)$ to a specific desired outcome $y'=h(\mathbf{x}')$.

Up to now, several algorithms for generating counterfactual explanations have been introduced. They are based on different principles, and for comprehensive surveys, see, e.g., \cite{Guidotti22,Verma2020}. Depending on the specific method, some properties of counterfactuals are expected to be met, such as \textit{validity} of the decision change, \textit{proximity} to the input instance, \textit{sparsity} of recommended changes, their \textit{actionability}, i.e., the counterfactual should not modify immutable features or violate monotonic constraints, and \textit{plausibility} of locating the counterfactual within a high-density region of the data, ensuring that the proposed counterfactuals are realistic and feasible within the context of the observed data distribution.

Many of these methods are inspired by the formulation of \citet{WachterMR17}, which proposed framing counterfactual explanations as an unconstrained optimization problem. For a prediction function $h$ and an input $\mathbf{x}_0 \in \mathbb{R}^d$, a counterfactual $\mathbf{x}' \in \mathbb{R}^d$ is computed by solving:
\begin{equation}
    \arg\min_{\mathbf{x}' \in \mathbb{R}^d} \ell(h(\mathbf{x}'), y') + C \cdot d(\mathbf{x}_0, \mathbf{x}') \text{.}
\label{eq:cf1}
\end{equation}
In this formulation, $\ell( \cdot , \cdot )$ represents a classification loss function, $d( \cdot , \cdot )$ is a penalty for deviation from the original input $\mathbf{x}_0$, and the term $C \geq 0$ serves as the regularization strength modifier.

An alternative approach~\cite{ArteltH19} frames counterfactual explanations as a constrained optimization problem. This perspective focuses on directly finding the minimal perturbation required to achieve the target prediction under the constraint that the model's prediction for the counterfactual instance meets the specified criterion. Mathematically, this is represented as:
\begin{equation}
\arg\min_{\mathbf{x}' \in \mathbb{R}^d} d(\mathbf{x}_0, \mathbf{x}') \quad \text{s.t.} \quad h(\mathbf{x}') = y' \text{.}
\label{eq:cf2}
\end{equation}

In our study, we want to pay special attention to the \textit{plausibility} of counterfactuals. Referring to arguments of \cite{Guidotti22}, a counterfactual is plausible if the feature values describing the example are coherent (sufficiently similar) with those present in the original data $X$. This means it should be located in sufficiently dense regions of original instances in $X$ from the target class. Plausibility helps in increasing users’ trust in the explanation: it would be hard to trust a counterfactual if it is a combination of features that are unrealistic with respect to existing examples.

In previous works, plausibility has often been verified by simple $k$-neighbourhood analysis of the counterfactual with respect to the original data~\cite{ShakhnarovichDI08, KeaneS20, SmythK22}. Few other approaches \cite{ArteltH20} model the conditional density in the target class and try to find the counterfactual example with the density value above the given threshold. Although the problem is quite well mathematically defined, the current methods apply simple approaches like kernel density estimators or a mixture of Gaussians to model conditional distributions that are difficult to apply for high-dimensional data. Moreover, the problem of estimating valid and plausible counterfactuals is defined as a complex constrained optimization problem with strict convexity assumptions \cite{ArteltH20}. Finally, the currently proposed methods, while providing valid counterfactuals, struggle to consistently produce observations that fulfill the plausibility criteria.

In this paper, we introduce \our{}: Probabilistically Plausibile Counterfactual Explanations using Normalizing Flows - a novel approach to estimate counterfactual explanations for differentiable classifiers tailored for tabular problems. It includes a novel, unconstrained formulation of the problem that enables direct estimation of the plausibility property - to the best of our knowledge, a characteristic previously not achieved in the literature. For that purpose, we design loss functions to satisfy both validity and plausibility constraints and minimize the distance to the original example in a balanced way (see an example in Fig. \ref{fig:moons}). Our approach incorporates plausibility in the probabilistic sense by targeting observations with a probability density exceeding a predefined threshold~\cite{ArteltH20}. Unlike existing methods limited to specific estimators of families of density functions, ours employs any differentiable conditional density model. Moreover, we postulate to utilize \textit{conditional normalizing flows} for density estimation \cite{rezende2015variational}, ensuring independence from specific parameterized distribution families while enabling direct calculation of density values for complex, high-dimensional data. Finally, \our{} leverages efficient batch processing utilizing gradient-based optimization techniques, leading to significant computational gains compared to previous methods.

To summarize, our contributions are as follows:
\begin{itemize}
    \item The formulation of counterfactual explanations within an unconstrained optimization framework employing direct optimization of plausibility and novel loss functions.
    \item The utilization of normalizing flows as density estimators to capture the complex high-dimensional data distribution effectively.
    \item The experimental evaluations demonstrating \our{}'s ability to efficiently generate high-quality, probabilistically plausible counterfactuals in high-dimensional tabular datasets for both binary and multiclass classification problems, outperforming existing reference methods.
\end{itemize}

\section{Related Works}
\subsection{Plausible Counterfactual Explanations}

The approaches for obtaining plausible counterfactual explanations are primarily categorized into \textit{endogenous} and \textit{exogenous} ones ~\cite{Guidotti22}. Endogenous counterfactuals are crafted using feature values from existing data instances, ensuring their naturally occurring status and grounding them in real-world contexts, thereby enhancing their plausibility. In contrast, exogenous counterfactuals are generated through methods such as interpolations or random data generation, which do not strictly rely on existing dataset features. While this offers greater flexibility, it does not inherently assure the plausibility of these counterfactuals, as they might represent feature combinations not found in actual data. 

\subsubsection{Endogenous Counterfactual Explanations}
Endogenous approaches to counterfactual explanations revolve around leveraging existing instances within the dataset to generate plausible counterfactuals. These methods, which include instance-based or case-based approaches, primarily utilize nearest neighbors' techniques to identify instances that closely resemble the input but yield different outcomes. 

Examples of endogenous methods include the Nearest-Neighbor Counterfactual Explainer (NNCE) \cite{ShakhnarovichDI08}, selecting similar yet outcome-divergent instances from the dataset as counterfactuals. The Case-Based Counterfactual Explainer (CBCE) \cite{KeaneS20} forms 'explanation cases' by pairing similar instances with contrasting outcomes, creating counterfactuals by merging features from these pairs. Extending this concept, the approach by \citet{SmythK22} adapts to k-nearest neighbors, utilizing multiple nearest candidates for generating counterfactuals. Feasible and Actionable Counterfactual Explanations (FACE) \cite{PoyiadziSSBF20} constructs a graph over data points, applying user-defined parameters to find actionable paths to desired outcomes. Lastly, PROPLACE \cite{JiangLLRT23} employs bi-level optimization and Mixed-Integer Linear Programming, generating robust counterfactuals from $\Delta$-robust nearest neighbors that closely align with data distribution and model robustness.

\subsubsection{Exogenous Counterfactual Explanations}
In the landscape of exogenous counterfactual explanations, methods generally involve introducing external modifications to original instances, diverging from reliance on existing instances and their features. These approaches utilize a range of computational techniques, such as autoencoders, linear programming, gradient-based methods, and generative models, to ensure that the resulting counterfactuals are plausible. 

Firstly, the Contrastive Explanation Method (CEM) \cite{DhurandharCLTTS18} innovates by adding perturbations to an instance and utilizing an autoencoder to verify the closeness of the modified instance to known data, ensuring plausibility. Meanwhile, the Diverse Coherent Explanations (DCE) \cite{Russell19} method leverages linear programming to create varied counterfactuals, with additional linear constraints to maintain both diversity and plausibility. Further, the Distribution-Aware Counterfactual Explanation (DACE) \cite{KanamoriTKA20} method incorporates the Mahalanobis distance and Local Outlier Factor (LOF) in its loss function, focusing on minimizing this distance while keeping a low LOF score to signify higher plausibility. The Diverse Counterfactual Explanations (DICE) \cite{MothilalST20} approach involves solving an optimization problem to generate multiple counterfactuals, with a specific emphasis on the diversity and actionability of these counterfactuals to determine their plausibility. Additionally, Counterfactual Explanations Guided by Prototypes (CEGP) \cite{LooverenK21} adopts a similar loss function to CEM but introduces a prototype-based loss term. This guides perturbations towards a counterfactual that aligns with the data distribution of a specific class, using the encoder of an autoencoder based on the average encoding of the nearest instances in the latent space with the same class label. 

Within the field of exogenous counterfactual explanations, a subcategory particularly relevant to our work utilizes deep generative models. Variational Autoencoders (VAEs) are exploited in methods like Example-Based Counterfactual (EBCF) \cite{MahajanTS19} and the approach by \citet{VerchevalP21}. EBCF incorporates known causal relationships into the VAE, promoting realistic counterfactuals. The method by \citet{VerchevalP21}  enables visual counterfactual generation through VAE-based latent space exploration. Generative Adversarial Networks (GANs) play a crucial role in the PCATTGAN approach~\cite{ArrietaS20}. It utilizes adversarial examples within a multi-objective optimization framework to create plausible counterfactuals, considering validity, minimality, and a notion of plausibility defined as human-understandable, non-automated changes. Diffusion models underpin methods proposed in \cite{JeanneretSJ22, AugustinBC022}. While these approaches specialize in visual counterfactual generation, their focus lies primarily on counterfactual sampling, not controlling plausibility via density-based optimization. Lastly, Normalizing Flow-based methods \cite{DomrowskiGMK22, DuongLX23} center on pinpointing counterfactuals within their latent spaces. These methods leverage the invertible nature of normalizing flows to explore counterfactual regions in the latent representation of the data.

All of the reference methods, except \citet{ArteltH20}, do not provide an explicit probabilistic formulations of plausibility. Compared to \citet{ArteltH20}, we propose an alternative problem formulation in unconstrained form with no prior constraints on the density model.

\section{Background}

In this work, we consider the problem formulation of probabilistically plausible counterfactual explanations introduced by \citet{ArteltH20}. This approach extends the problem formulation given by eq. \eqref{eq:cf2} by adding a target-specific density constraint to enforce the plausibility of counterfactuals using a probabilistic framework. The constrained optimization problem is formulated as follows:
\begin{subequations}
\begin{align}
    \arg\min_{\mathbf{x}' \in \mathbb{R}^d} d(\mathbf{x}_0, \mathbf{x}')\\
    \text{s.t.} \quad h(\mathbf{x}') = y'\\
    % \hat{p}_y(\mathbf{x}') \geq \delta
    \delta \leq p(\mathbf{x'}|y'), \label{eq:prob_plaus}
\end{align}
\label{eq:problem_statement}
\end{subequations}
where  $p(\mathbf{x}'|y')$ denotes conditional probability of the counterfactual explanation $\mathbf{x}'$ under desired target class value $y'$ and $\delta$ represents the density threshold. 

This approach's crucial aspect is finding the proper model to represent the conditional density function $p(\mathbf{x}|y)$. Typically, kernel density estimators (KDEs) are used to model conditional densities, but the use of non-linear kernels results in the  highly non-convex optimization problem formulation. Gaussian Mixture Model (GMM) can be applied alternatively, but convexity constraints are still not satisfied. To facilitate the desired optimization process, the authors of \citet{ArteltH20} propose to approximate the density value $p(\mathbf{x}'|y')$ using a component-wise maximum of GMM components:

\begin{equation}
    \hat{p}_G(\mathbf{x}'|y') = \max_{j} \Bigl( \pi_{j,y'} \mathcal{N}(\mathbf{x}' | \boldsymbol{\mu}_{j,y'}, \mathbf{\Sigma}_{j,y'}) \Bigr),
\end{equation}
where $\boldsymbol{\mu}_{j,y'}$, $\mathbf{\Sigma}_{j,y'}$ and $\pi_{j,y'}$ are means, covariances and prior values for component $j$ considering class $y$. 

This approximation is transformed into a convex quadratic constraint for each GMM component $j$, resulting in the following formula:
\begin{equation}
    (\mathbf{x}' - \boldsymbol{\mu}_{j,y'})^T \mathbf{\Sigma}_{j,y'} (\mathbf{x}' - \boldsymbol{\mu}_{j,y'}) + c_{j} \leq \delta',
\end{equation}
where $c_j$ is constant from the Gaussian normalization factor and $\delta'=-2 \log{\delta}$. 

For each component $j$, the optimization problem is solved, resulting in a set of convex programs - one for each component. This step is crucial because knowing beforehand which component will produce a feasible and plausible counterfactual is impossible. Finally, the counterfactual $\mathbf{x'}$ that yields the smallest value for the objective function is selected.

However, this approach has few limitations. First, the number of components should be predefined for each class. Second, the family of parametrized distributions limits the ability to adjust to a data distribution. Third, the approach is difficult to be applied to high-dimensional data due to the Gaussian components.

In order to cope with the listed limitations, we postulate to model conditional density function $p(\mathbf{x}|y)$ using the normalizing flows \cite{rezende2015variational}. This group of models can adjust to very complex, high-dimensional data distributions, which allows for calculating the density value from the change-of-variable formula. Moreover, we propose an alternative unconstrained problem formulation that allows solving using a gradient-based approach for any differentiable representation of conditional distribution $p(\mathbf{x}|y)$.

\section{Method}
This section introduces a novel approach to the problem of plausible counterfactual explanation formulated by eq. \eqref{eq:problem_statement}. First, we reformulate the problem of calculating counterfactuals as unconstrained optimization suitable for direct, gradient-based optimization. Next, we show how to train the flow model to estimate the class-conditional distributions. Finally, we show how the counterfactuals can be efficiently estimated using a gradient-based approach.  

\subsection{Unconstrained Probabilistically Plausible Counterfactual Explanations}

We consider a binary classification problem, $y \in \{0,1\}$. However, our considerations can be easily extended to the multiclass case. Further, we consider a discriminative differentiable model (e.g., Logistic Regression or MLP) $p_d(y|\mathbf{x})$ and reformulate the validity constraint $h(\mathbf{x}') = y'$  as $p_d(y'|\mathbf{x}')\geq 0.5 + \epsilon$, where $\epsilon \to  0$, practically represented as small enough value close to $0$. 

We postulate the following unconstrained optimization problem: 
\begin{equation}
    \arg\min_{\mathbf{x}'  \in \mathbb{R}^d} d(\mathbf{x}_0, \mathbf{x}') + \lambda \cdot \biggl( \ell_{v}(\mathbf{x}', y') + \ell_{p}(\mathbf{x}', y') \biggr)  \text{,}
\label{eq:our_cf1}
\end{equation}
where $\lambda=\infty$, practically, is large enough. 

The loss $\ell_{v}(\mathbf{x}', y')$ component controls the validity constraint and is defined as follows:

\begin{equation}
\ell_{v}(\mathbf{x}', y')=\max \Bigl(0.5 + \epsilon - p_d(y'|\mathbf{x}'), 0 \Bigr).
\end{equation}

The Binary Cross Entropy (BCE) criterion can be used alternatively. However, using such criteria enforces $100\%$ confidence of the discriminative model, while our approach aims at achieving the current classification accuracy with the margin controlled with the $\epsilon$ parameter. While using our criterion, the model can focus more on producing closer and more plausible counterfactuals, which we show in ablation studies. 

Additionally, we extend the validity loss component to the multiclass scenario in the following way:
\begin{equation}
\ell_{v}(\mathbf{x}', y')=\max \Bigl(\max_{y \neq y'}p_d(y|\mathbf{x}') + \epsilon - p_d(y'|\mathbf{x}'), 0 \Bigr),
\end{equation}
where we replace the $0.5$ threshold value with the highest probability value returned by the discriminative model, excluding the value for target class $y'$. This guarantees that $p_d(y|\mathbf{x}')$ will be higher than the most probable class among the remaining classes by the $\epsilon$ margin.

The loss component $\ell_{p}(\mathbf{x}', y')$ controls probabilistic plausibility constraint ($\delta \leq p(\mathbf{x'}|y')$) and is defined as:
\begin{equation}
\ell_{p}(\mathbf{x}', y') = \max \Bigl(\delta - p(\mathbf{x'}|y') , 0 \Bigr),
\end{equation}
where $\delta$ is the density threshold calculated in the same way as in \cite{ArteltH20}, i.e., by utilizing the median of the training dataset. The conditional distribution $p(\mathbf{x}|y)$ can be represented by any differentiable model (e.g., Mixture of Gaussians, KDE). In this work, we postulate to model the distribution using conditional normalizing flow due to the flexibility and ability to adjust to multidimensional complex distributions. Thanks to the unconstrained problem formulation given by eq. \eqref{eq:our_cf1} and differentiation assumption for the models, the counterfactuals can be easily calculated using a gradient-based approach.

\subsection{Probabistically Plausible Counterfactual Explanations via Normalizing Flow-based Density Estimation}
\label{sec:method_nfs}

KDE or GMMs can be used to model the conditional distributions. However, those models have limited modeling capabilities due to the parametrized (usually Gaussian) form of $p(\mathbf{x}|y)$ or the inability to model high-dimensional data (KDE). Therefore, in this work, we postulate the use of a conditional normalizing flow model \cite{rezende2015variational} to estimate the density for the joint distribution of the attributes for each class.

Normalizing Flows have surged in popularity within generative models due to their adaptability and the simplicity of training via direct negative log-likelihood (NLL) optimization. Their adaptability stems from the change-of-variable technique, which transforms a latent variable $\mathbf{z}$ with a known prior distribution $p(\mathbf{z})$ into an observed space variable $\mathbf{x}$ with an unknown distribution. This transformation occurs through a sequence of invertible (parametric) functions: $\mathbf{x}=\mathbf{f}_K \circ \dots \circ \mathbf{f}_1(\mathbf{z},y)$. Assuming a known prior $p(\mathbf{z})$ for $\mathbf{z}$, the conditional log-likelihood for $\mathbf{x}$ is expressed as:
\begin{equation}
\log \hat{p}_F(\mathbf{x}|y) = \log p(\mathbf{z}) - \sum_{k=1}^K \log  \left| \det \frac{\partial \mathbf{f}_k}{\partial \mathbf{z}_{k-1}} \right|,
\label{eq:flow_p_x_y}
\end{equation}
where $\mathbf{z} = \mathbf{f}_1^{-1}\circ \dots \circ \mathbf{f}_K^{-1} (\mathbf{x}, y)$ is a result of the invertible mapping. The biggest challenge in normalizing flows is the choice of the invertible functions $\mathbf{f}_K, \dots, \mathbf{f}_1$. Several solutions have been proposed in the literature to address this issue with notable approaches, including NICE \cite{DinhKB14}, RealNVP \cite{DinhSB17}, and MAF \cite{PapamakariosMP17}.

For a given training set $\mathcal{D}=\{ (\mathbf{x}_n, y_n)\}_{n=1}^N$ we simply train the conditional normalizing flow by minimizing negative log-likelihood:

\begin{equation}
Q = - \sum_{n=1}^N \log \hat{p}_F(\mathbf{x}_n|y_n),
\end{equation}
where $\log \hat{p}_F(\mathbf{x}_n|y_n)$ is defined by eq. \eqref{eq:flow_p_x_y}. The model is trained using a gradient-based approach applied to the flow parameters stored in $\mathbf{f}_k$ functions. 

% For some practical counterfactuals applications, where the labels are not given, we can use a simple distillation technique to assign the labels for unsupervised data $\mathcal{X}=\{ \mathbf{x}_n \}_{n=1}^N$ using discriminative model $p_d(y|\mathbf{x})$ as an oracle. We create labeled dataset $D_d=\{(\mathbf{x}_n,\hat{y}_n)\}_{n=1}^N$, where $\hat{y}_n = \arg \max_{y}p_d(y|\mathbf{x}_n)$ and utilize it to train the conditional normalizing flow. 

\subsection{Estimating Counterfactuals}

For a trained conditional normalizing flow, the counterfactual explanation can be easily calculated simply by optimizing the criterion given by eq. \eqref{eq:our_cf1}. The parameters of the flow model are frozen, and $\mathbf{x}'$ is optimized using the gradient-based procedure, starting from the point $\mathbf{x}_0$. To enhance the efficiency of our method, we have incorporated batch processing capabilities, allowing for the simultaneous calculation of multiple counterfactual explanations. This is achieved by aggregating instances and employing an average aggregation for loss calculation. Such a feature is notably absent in the other approaches compared to this study, providing our method with a distinct computational advantage.

% We can consider the second scenario, for which we have the discriminative trained model $p(y|x)$ (logistic regression, MLP) and unconditioned generative model $p(\mathbf{x})$ (flow trained using unlabelled data) in this case, we can simply substitute $p(\mathbf{x}|y)$:

% \begin{equation}
% p(x|y) = \frac{p(y|x)p(x)}{p(y)}
% \end{equation}

% \todo{We might take a look at \cite{ArteltH20} regarding why we can drop classification constraint.}

\section{Experiments}
\begin{table*}[t]
\centering
\caption{Comparative Results of Probabilistically Plausible Counterfactual Explanation Methods. We contrast the performance of \our{} method with  Artelt \& Hammer \cite{ArteltH20} and other methods across Logistic Regression (LR) classifier. The results demonstrate our method's consistently valid and probabilistically plausible results and its ability to produce counterfactuals even in complex scenarios like high-dimensional data.}
\label{tab:ours_vs_all_lr}

\begin{center}
\begin{sc}
\begin{scriptsize}
\begin{tabular}{l|l|rrrrrrrrr}
\toprule
Dataset & Method & Coverage $\uparrow$ & Validity $\uparrow$ & Prob. Plaus. $\uparrow$ & LOF & IsoForest & Log Dens. $\uparrow$ & L1 $\downarrow$ & L2 $\downarrow$ & Time $\downarrow$ \\
\midrule
\multirow{6}{*}{Moons} & CBCE & \textbf{1.00} & \textbf{1.00} & 0.10 & 1.06 & 0.03 & -5.81 & 0.62 & 0.48 & \textbf{0.07} s\\
& CEGP & \textbf{1.00} & \textbf{1.00} & 0.09 & 1.36 & 0.00 & -6.66 & 0.36 & \textbf{0.28} & 904.11 s\\
& CEM & \textbf{1.00} & \textbf{1.00} & 0.14 & 2.03 & -0.07 & -10.09 & 0.55 & 0.50 & 211.56 s\\
& WACH & 0.98 & \textbf{1.00} & 0.11 & 1.55 & -0.01 & -6.34 & 0.49 & 0.36 & 198.29 s\\
\cmidrule{2-11}
& ARTELT & \textbf{1.00} & \textbf{1.00} & 0.08 & 1.53 & -0.03 & -8.74 & \textbf{0.32} & 0.32 & 4.15 s\\
& \our{} & \textbf{1.00} & \textbf{1.00} & \textbf{1.00} & 1.01 & 0.04 & \textbf{1.69} & 0.45 & 0.36 & 1.85 s\\
\midrule
\multirow{6}{*}{Law} & CBCE & \textbf{1.00} & \textbf{1.00} & 0.49 & 1.05 & 0.04 & 1.28 & 0.61 & 0.40 & \textbf{0.23} s\\
& CEGP & \textbf{1.00} & \textbf{1.00} & 0.49 & 1.07 & 0.04 & 1.08 & 0.23 & 0.18 & 1973.76 s\\
& CEM & \textbf{1.00} & \textbf{1.00} & 0.26 & 1.26 & -0.02 & -0.56 & 0.33 & 0.31 & 368.10 s\\
& WACH & \textbf{1.00} & \textbf{1.00} & 0.39 & 1.30 & -0.01 & -0.29 & 0.45 & 0.35 & 359.00 s\\
\cmidrule{2-11}
& ARTELT & \textbf{1.00} & \textbf{1.00} & 0.40 & 1.12 & 0.02 & 0.54 & \textbf{0.20} & \textbf{0.20} & 4.02 s\\
& \our{} & \textbf{1.00} & \textbf{1.00} & \textbf{1.00} & 1.03 & 0.07 & \textbf{2.05} & 0.37 & 0.23 & 2.42 s\\
\midrule
\multirow{6}{*}{Audit} & CBCE & \textbf{1.00} & \textbf{1.00} & 0.79 & 11.70 & 0.14 & \textbf{54.97} & 2.55 & 1.24 & \textbf{0.04} s\\
& CEGP & 0.97 & \textbf{1.00} & 0.02 & 6.08$\cdot\text{10}^7$ & 0.02 & 8.09 & 1.56 & 0.57 & 561.04 s\\
& CEM & 0.52 & \textbf{1.00} & 0.00 & 8.28$\cdot\text{10}^6$ & -0.04 & 20.84 & 1.20 & \textbf{0.37} & 105.92 s\\
& WACH & \textbf{0.99} & \textbf{1.00} & 0.02 & 1.42$\cdot\text{10}^8$ & 0.06 & -40.34 & 1.78 & 0.80 & 101.27 s\\
\cmidrule{2-11}
& ARTELT & 0.60 & 0.97 & 0.00 & 4.09$\cdot\text{10}^8$ & 0.10 & -3585.76 & \textbf{0.90} & 0.88 & 43.84 s\\
& \our{} & \textbf{1.00} & \textbf{0.99} & \textbf{0.99} & 4.25$\cdot\text{10}^7$ & 0.08 & 51.64 & 2.04 & 0.79 & 7.01 s\\
\midrule
\multirow{6}{*}{Heloc} & CBCE & \textbf{1.00} & \textbf{1.00} & 0.54 & 1.10 & 0.07 & 28.01 & 2.84 & 0.82 & \textbf{5.71} s\\
& CEGP & \textbf{1.00} & \textbf{1.00} & 0.29 & 3.50$\cdot\text{10}^7$ & 0.04 & 24.75 & \textbf{0.26} & \textbf{0.10} & 9654.60 s\\
& CEM & \textbf{1.00} & \textbf{1.00} & 0.07 & 2.50$\cdot\text{10}^8$ & 0.02 & 12.37 & 0.35 & 0.20 & 1639.16 s\\
& WACH & \textbf{1.00} & \textbf{1.00} & 0.00 & 2.65$\cdot\text{10}^8$ & 0.03 & -15.09 & 0.74 & 0.37 & 1600.28 s\\
\cmidrule{2-11}
& ARTELT & 0.00 & - & - & - & - & - & - & - & - s\\
& \our{} & \textbf{1.00} & \textbf{1.00} & \textbf{1.00} & 6.47$\cdot\text{10}^7$ & 0.07 & \textbf{32.42} & 0.90 & 0.23 & 12.44 s\\
\midrule
\multirow{6}{*}{Blobs} & CBCE & \textbf{1.00} & \textbf{1.00} & 0.27 & 1.02 & 0.03 & -35.52 & 0.95 & 0.72 & \textbf{0.13} s\\
& CEGP & \textbf{1.00} & \textbf{1.00} & 0.00 & 2.43 & -0.07 & -9.08 & \textbf{0.30} & \textbf{0.25} & 1295.36 s\\
& CEM & 0.96 & \textbf{1.00} & 0.00 & 3.51 & -0.12 & -14.95 & 0.46 & 0.45 & 512.56 s\\
& WACH & \textbf{1.00} & \textbf{1.00} & 0.04 & 2.24 & -0.06 & -9.52 & 0.51 & 0.38 & 441.59 s\\
\cmidrule{2-11}
& ARTELT & \textbf{1.00} & \textbf{1.00} & 0.00 & 2.11 & -0.07 & -3.51 & 0.39 & 0.33 & 6.62 s\\
& \our{} & \textbf{1.00} & \textbf{1.00} & \textbf{1.00} & 1.01 & 0.04 & \textbf{3.00} & 0.69 & 0.50 & 3.22 s\\
\midrule
\multirow{6}{*}{Digits} & CBCE & \textbf{1.00} & \textbf{1.00} & 0.18 & 1.02 & 0.04 & 23.72 & 16.28 & 3.09 & \textbf{0.51} s\\
& CEGP & \textbf{1.00} & \textbf{1.00} & 0.11 & 1.09 & 0.01 & -0.39 & 2.53 & \textbf{0.63} & 1945.67 s\\
& CEM & \textbf{1.00} & 0.98 & 0.01 & 1.23 & -0.03 & -86.77 & 5.28 & 1.38 & 852.05 s\\
& WACH & \textbf{1.00} & \textbf{1.00} & 0.08 & 1.20 & 0.00 & -34.97 & \textbf{2.47} & 1.20 & 651.00 s\\
\cmidrule{2-11}
& ARTELT & 0.80 & 0.93 & 0.04 & 1.69 & 0.01 & -54.72 & 3.30 & 2.43 & 238.28 s\\
& \our{} & \textbf{1.00} & \textbf{1.00} & \textbf{1.00} & 1.12 & 0.03 & \textbf{44.42} & 8.27 & 1.33 & 8.68 s\\
\midrule
\multirow{6}{*}{Wine} & CBCE & \textbf{1.00} & \textbf{1.00} & 0.37 & 1.06 & 0.05 & 2.13 & 3.38 & 1.12 & \textbf{0.01} s\\
& CEGP & \textbf{1.00} & \textbf{1.00} & 0.01 & 1.08 & 0.05 & -0.15 & 0.82 & 0.32 & 191.09 s\\
& CEM & \textbf{1.00} & \textbf{1.00} & 0.00 & 1.35 & -0.02 & -12.94 & 1.20 & 0.63 & 81.33 s\\
& WACH & \textbf{1.00} & \textbf{1.00} & 0.01 & 1.27 & 0.00 & -9.41 & 1.57 & 0.78 & 50.74 s\\
\cmidrule{2-11}
& ARTELT & \textbf{1.00} & 0.97 & 0.01 & 1.33 & 0.02 & -11.73 & 0.68 & 0.65 & 0.96 s\\
& \our{} & \textbf{1.00} & \textbf{1.00} & \textbf{1.00} & 1.01 & 0.09 & \textbf{9.72} & 1.65 & 0.53 & 2.03 s\\
\bottomrule
\end{tabular}
\end{scriptsize}
\end{sc}
\end{center}
\end{table*}

In this section, we aim to demonstrate and validate our counterfactual explanation method through a series of experiments. Initially, we illustrate our method's intuition with the Moons dataset and Logistic Regression model. Next, we compare our approach against the only reference method in a probabilistically plausible CFs area - \citet{ArteltH20}, as well as other established CF methods. This comparison focuses on the impact of plausibility on proximity metrics and time efficiency. Lastly, we conduct broader comparisons using other classifier models: Logistic Regression (LR), Multilayer Perceptron (MLP), and Neural Oblivious Decision Ensembles (NODE)~\cite{PopovMB20}. The code for these experiments is publicly released on GitHub\footnote{\url{https://github.com/ofurman/counterfactuals}}.

\paragraph{Datasets}
To evaluate \our{}'s effectiveness, we conducted experiments on seven numerical-only tabular datasets.  Four datasets (Law, Heloc, Moons, and Audit) represent binary classification problems, whereas the first two datasets (Law and Heloc) are commonly used benchmarks for counterfactual explanation tasks. The remaining three datasets (Blobs, Digits, and Wine) address multiclass classification problems. Detailed descriptions of these datasets are available in the Appendix \ref{apx:datasets}~\cite{abs-2405-17640}. Overall, they represent broad diversity in sample sizes (up to approximately 10.000), number of variables (up to 64), and number of classes (up to 10). For preprocessing purposes, we implemented two key steps to prepare the datasets. First, we addressed class imbalance by downsampling the majority class to match the size of the minority class. Second, we normalized all features across the datasets to a $[0, 1]$ range, enabling consistent scale and comparability among features. Thirdly, to ensure robust method evaluation, we employed stratified 5-fold cross-validation on each dataset. Finally, for clarity, the main manuscript reports average values, while the appendix~\cite{abs-2405-17640} includes standard deviation for detailed analysis.
% PW Comment: Too detailed. \of{This involved randomly splitting each dataset into five equally sized folds. The experiments were then performed on four folds and evaluated on the remaining unseen fold. This process was repeated five times, ensuring all data points contribute to both training and testing phases.}

\paragraph{Classification Models}
For the experiments, we include Logistic Regression (LR), 3-layer Multilayer Perceptron (MLP), and Neural Oblivious Decision Ensemble (NODE) catering to both linear and non-linear scenarios. LR aligns with linear assumptions prevalent in some baseline methods, MLP allows for the assessment of behaviors in non-linear model contexts, and NODE stands as an example of a complex ensemble of neural decision trees. This triple-model approach facilitates a thorough evaluation across varied model complexities. Crucially, all models are differentiable, which is essential in the context of our method.

\paragraph{Experiments Details}
For every combination of the classification model and dataset, we trained both the classification model and a Normalizing Flow as the density estimator, following the approach detailed in Section \ref{sec:method_nfs}. We opted for the Masked Autoregressive Flow (MAF) architecture \cite{PapamakariosMP17} as our choice for the Normalizing Flow. This decision was based on experimental findings indicating MAF's superior performance in accurately fitting data distributions. For a deeper analysis of these results, including in-depth model performance metrics like accuracy, please refer to the Appendix~\cite{abs-2405-17640}. See Section \ref{sec:density_estimator} for a detailed exploration and Tab. \ref{tab:dataset_description} for specific performance figures. The final step involved generating counterfactual explanations for the entire set of test samples.

\paragraph{Reference Methods}
Our analysis includes several significant baselines, each selected for its relevance to the field. We first consider the method developed by Artelt and Hammer \cite{ArteltH20}, notable for its focus on probabilistically plausible counterfactuals. Additionally, we evaluate the approach by \citet{WachterMR17}, widely recognized as a foundational baseline in counterfactual explanations research. To provide both endogenous and exogenous counterfactual explanations, we compare three methods: Case-Based Counterfactual Explainer (CBCE) \cite{KeaneS20}, Contrastive Explanation Method (CEM) \cite{DhurandharCLTTS18}, and Counterfactual Explanations Guided by Prototypes (CEGP) \cite{LooverenK21}.
% \pw{Finally, we include model by \citet{DomrowskiGMK22} which also utilizes Normalizing Flows.} It's worth noting that both the Wachter and CEGP methods were acknowledged for their performance in implausibility metrics in the comprehensive review by \citet{Guidotti22}.

\paragraph{Metrics}
Following related works, we chose a comprehensive set of metrics to assess the performance of counterfactual explanation methods. We include two success metrics: \textit{coverage}, evaluating the method's ability to generate explanations across all instances, and \textit{validity}, assessing the efficacy of counterfactuals in altering the model's decision. In terms of proximity, we measure the \textit{L1} and \textit{L2} distances to quantify the closeness between original instances and their counterfactuals. We evaluate plausibility using a combination of metrics. First of all, we measure the \textit{Local Outlier Factor (LOF)} score, which, when significantly greater than 1, indicates an outlier, with values closer to 1 suggesting normalcy, highlighting anomalies through local density deviations. Secondly, we utilize \textit{Isolation Forest}, which assigns scores between -0.5 and 0.5, with values approaching -0.5 identifying anomalies due to the ease of isolation and scores above 0 indicating normal observations. We further access counterfactuals using \textit{probabilistic plausibility} metric, the proportion of CFs meeting the criterion defined in Eq. \ref{eq:prob_plaus}. Moreover, we calculate \textit{log density}, which gauges the logarithmic probability density of counterfactuals under the target class, with higher values indicating greater plausibility. Finally, we calculate \textit{time} metric representing time in seconds needed for the method to process the whole test dataset.

\subsection{Method Intuition via Toy Example}
In our illustrative example, we present the counterfactual generation process using the Moons dataset under a Logistic Regression model, as depicted in Figure \ref{fig:moons}. The initial observation is represented by a black dot, with intermediary observations during the optimization process (after every 150 iteration steps) shown as orange dots and the final counterfactual outcome marked by a red dot. The probability distributions are indicated by contour lines, with the filled red contour denoting the region exceeding the desired density threshold. The blue line illustrates the decision boundary of the classifier. This visualization effectively demonstrates how our method navigates toward the target classification and probabilistic plausibility regions, adjusting its trajectory to surpass the classifier's decision boundary by a predefined margin $\epsilon$ upon achieving the required density level.

\begin{table*}[ht]
\centering
\caption{Analysis of Counterfactual Methods Across Classification Models. We offer a detailed comparison of our method and other well-established reference methods across two classification models: a 3-layer Multilayer Perceptron (MLP), and a Neural Oblivious Decision Ensemble (NODE). The results emphasize the efficacy of our method in producing valid and plausible counterfactuals across various models, including those that are deeper and more complex.}
\label{tab:ours_vs_references}
\begin{center}
\begin{sc}
\begin{scriptsize}
\begin{tabular}{l|l|rrrrrrrrr}
\toprule
Dataset & Method & Cov. $\uparrow$ & Val. $\uparrow$ & Prob. Plaus. $\uparrow$ & LOF & IsoForest & Log Dens. $\uparrow$ & L1 $\downarrow$ & L2 $\downarrow$ & Time $\downarrow$ \\
\midrule
\multicolumn{11}{c}{MLP} \\
\midrule
\multirow{6}{*}{Heloc} & CBCE & \textbf{1.00} & \textbf{0.94} & 0.54 & 1.09 & 0.08 & 28.85 & 2.87 & 0.82 & \textbf{6.47} s\\
 & CEGP & 0.94 & 0.63 & 0.05 & 4.15$\cdot\text{10}^8$ & 0.01 & -3.28 & 1.25 & 0.43 & 31309.33 s\\
 & CEM & \textbf{1.00} & 0.86 & 0.01 & 7.71$\cdot\text{10}^8$ & -0.01 & -89.39 & 1.32 & 0.58 & 6938.45 s\\
 & WACH & \textbf{0.99} & 0.81 & 0.00 & 1.34$\cdot\text{10}^8$ & -0.06 & -161.68 & 3.11 & 0.90 & 23392.40 s\\
 & ARTELT & - & - & - & - & - & - & - & - & - s\\
 & \our{} & \textbf{1.00} & 0.92 & \textbf{1.00} & 1.42$\cdot\text{10}^8$ & 0.07 & \textbf{32.07} & \textbf{1.18} & \textbf{0.31} & 25.32 s\\
\midrule
\multirow{6}{*}{Digits} & CBCE & \textbf{1.00} & \textbf{1.00} & 0.18 & 1.02 & 0.04 & 23.66 & 16.29 & 3.09 & \textbf{0.54} s\\
 & CEGP & 0.95 & 0.46 & 0.02 & 1.24 & -0.02 & -138.62 & 6.39 & \textbf{1.42} & 2523.28 s\\
 & CEM & \textbf{1.00} & 0.42 & 0.01 & 1.44 & -0.06 & -481.57 & \textbf{6.34} & 1.76 & 1260.54 s\\
 & WACH & \textbf{1.00} & 0.72 & 0.00 & 1.50 & -0.07 & -516.44 & 11.04 & 2.13 & 3342.38 s\\
 & ARTELT & - & - & - & - & - & - & - & - & - s\\
 & \our{} & \textbf{1.00} & \textbf{1.00} & \textbf{0.98} & 1.13 & 0.03 & \textbf{43.87} & 8.78 & \textbf{1.42} & 25.09 s\\
\midrule
\multicolumn{11}{c}{NODE} \\
\midrule
\multirow{5}{*}{Heloc} & CBCE & \textbf{1.00} & \textbf{1.00} & 0.55 & 1.09 & 0.08 & 28.88 & 2.85 & 0.82 & \textbf{17.53} s\\
& CEM & 0.94 & \textbf{1.00} & 0.10 & 1.35 & 0.05 & 9.00 & \textbf{0.47} & 0.29 & 14772.66 s\\
 & WACH & 0.96 & \textbf{1.00} & 0.10 & 2.12$\cdot\text{10}^8$ & 0.05 & 10.75 & 0.85 & 0.36 & 37254.33 s\\
 & ARTELT & - & - & - & - & - & - & - & - & - s\\
 & \our{} & \textbf{1.00} & 0.94 & \textbf{1.00} & 1.08 & 0.09 & \textbf{31.85} & 1.02 & \textbf{0.28} & 126.05 s\\
\midrule
\multirow{5}{*}{Digits} & CBCE & \textbf{1.00} & \textbf{1.00} & 0.18 & 1.02 & 0.04 & 24.00 & 16.27 & 3.09 & \textbf{3.12} s\\
& CEM & \textbf{1.00} & \textbf{1.00} & 0.03 & 1.32 & -0.02 & -39.458 & 4.07 & 1.44 & 5451.835 s\\
 & WACH & \textbf{1.00} & \textbf{1.00} & 0.16 & 1.12 & 0.02 & 7.02 & \textbf{2.93} & \textbf{1.13} & 15376.44 s\\
 & ARTELT & - & - & - & - & - & - & - & - & - s\\
 & \our{} & \textbf{1.00} & \textbf{1.00} & \textbf{1.00} & 1.15 & 0.02 & \textbf{43.97} & 7.76 & 1.36 & 69.45 s\\
\bottomrule
\end{tabular}
\end{scriptsize}
\end{sc}
\end{center}
\end{table*}

\subsection{Probabilistically Plausibile Counterfactual Explanations Methods Comparison}
In this section, we conduct a focused comparison of our approach, \our{}, against the method by \citet{ArteltH20}, which is the primary reference in the realm of probabilistically plausible counterfactual explanations. For that purpose, we utilize the datasets, metrics, and classifiers described in the previous section. The evaluation is centered on assessing and validating the accuracy of both methods in generating counterfactuals, their plausibility, and their proximity to original instances.

The results are presented in Tab. \ref{tab:ours_vs_all_lr}. Firstly, we can observe that our method always returns the results that are probabilistically plausible. That is not the case for Artelt's method, which struggles in high-dimensional datasets like Heloc (23 dimensions) or Digits (64 dimensions), doesn't support non-linear classifiers like MLPs, and wasn't able to consistently fulfill the probabilistic plausibility criterion. Secondly, in terms of distances, Artelt's method returns better results, which is expected due to the trade-off between distance and plausibility, i.e., the more plausible observations, the farther away they usually are. However, the results are not clearly worse, especially in terms of L2 distance, meaning \our{} can balance both desired properties of counterfactuals. Thirdly, the log density values of the observations produced by \our{} method are significantly better. Fourthly, our analysis using Local Outlier Factor (LOF) and Isolation Forest (IsoForest) metrics indicates that our methods generate inliers (except for Audit and Heloc, where almost all methods struggle to obtain reasonable values of LOF), whereas Artelt's method underperforms and can sometimes result in outliers. Fifthly, our method turned out to be significantly faster, with the speed up around x2-10 on relatively small datasets. Finally, our method was almost always able to produce valid counterfactual explanations for MLP and NODE, contrary to Artelt (see results in Tab. \ref{tab:ours_vs_references} and detailed results in Tab. \ref{tab:ours_vs_all_cv_mlp} and \ref{tab:ours_vs_all_cv_node} in Appendix~\cite{abs-2405-17640}). It's worth mentioning that \our{} almost always returned probabilistically plausible observations, which, in case of non-valid observations, might still be valuable insight for the final user, contrary to the lack of a response at all.

% \pw{The described non-valid yet plausible counterfactual refers to one that remains below the classification threshold yet resides within the designated plausibility region. For instance, in Figure 1, consider the segment above the decision boundary but within the red-highlighted probabilistically plausible zone, which delineates an area of such CFs. This area could be seen as where the classifier's inaccuracies lie, given that the plausibility zone is defined based on true labels. The differences may occur due to not perfect adjustments to the training data considering both models that are trained - they are not perfect, just modeling. While such instances may not provide actionable insights for the model's end-users, they are invaluable for developers, offering a lens through which to identify and address the classifier's vulnerabilities. Finally, the adjustment may occur if we use the relabelling using the classification model to train the flow would be applied.}

\subsection{Counterfactual Explanations Methods Comparison}
In this comparative analysis, we evaluate our method against well-established reference methods, with a particular focus on the impact of integrating probabilistically plausible conditions into the optimization process. Our primary objective is to assess our method's performance in terms of validity, plausibility, proximity metrics, and processing efficiency. We also explore whether methods not specifically designed for plausibility can still produce plausible counterfactuals across various classifiers such as Logistic Regression (LR), Multilayer Perceptron (MLP), and Neural Oblivious Decision Ensembles (NODE).

Results presented in Tab. \ref{tab:ours_vs_all_lr} and Tab. \ref{tab:ours_vs_references} indicate that our model excels in validity, plausibility (considering both probabilistic formulation and outlier metrics), and processing times while maintaining reasonable distances compared to competing approaches across all datasets and classification methods. Specifically, Tab. \ref{tab:ours_vs_references} presents the evaluation results for two selected high-dimensional datasets (one for binary classification problem and one for multiclass problem) using two advanced classifiers, demonstrating that our method consistently produces valid results not only with a shallow model, such as LR but also with deeper models, including MLP and NODE. In contrast, the majority of existing methods encounter difficulties in producing valid counterfactual explanations for the Multilayer Perceptron. We conducted a comprehensive evaluation utilizing all methods and datasets mentioned earlier, applying three different classifiers. Detailed outcomes are presented in Appendix \ref{apx:additional_results}~\cite{abs-2405-17640}. Particularly, results for Logistic Regression are shown in Tab. \ref{tab:ours_vs_all_cv_lr}, while findings for the MLP and NODE classifiers are detailed in Tab. \ref{tab:ours_vs_all_cv_mlp} and \ref{tab:ours_vs_all_cv_node}, respectively.

Furthermore, our hypothesis that reference methods could inadvertently yield plausible outcomes without targeted optimization was not confirmed. In terms of proximity, CEGP achieves the most favorable outcomes, with our method typically ranking closely behind. This demonstrates our method's effectiveness in balancing proximity and plausibility constraints. Notably, our method's computational time efficiency closely parallels the CBCE method, which does not involve an optimization process. This efficiency is due to our batching strategy, which processes all datasets collectively, as opposed to the case-by-case optimization typical of other methods. Summarizing, our method generates probabilistically plausible counterfactuals with exceptional efficiency and minimal compromise on proximity. Its ability to process high-dimensional data quickly makes it ideal for resource-constrained, real-world applications.

\section{Method Analysis}
In this section, we delve into the analysis of two pivotal components of our proposed method: the loss function and the regularization hyperparameter $\lambda$. Adhering to the experimental framework established in the earlier sections, these studies are conducted specifically using the Logistic Regression model. Our focus is on evaluating the impact of these elements on the method's overall performance and efficacy.

\subsection{Loss Function Ablation Study}

\begin{table}[t]
\centering
\caption{Ablation Study on Loss Function Selection.}
%  We compare the impact of using our proposed loss function versus Binary Cross Entropy (BCE) on counterfactual explanation metrics across various datasets. The results demonstrate that our loss function is more effective in producing closer and more plausible counterfactuals.
\label{tab:ablation_loss}

\begin{center}
\begin{sc}
\begin{scriptsize}
\begin{tabular}{l|l|rrrrrr}
\toprule
Dataset & Loss & Cov. & Val. & PP & L1 & L2 & LD \\
\midrule
\multirow{2}{*}{Moons} & Ours & \textbf{1.00} & \textbf{1.00} & \textbf{1.00} & \textbf{0.45} & \textbf{0.36} & 1.69 \\
 & BCE & \textbf{1.00} & \textbf{1.00} & \textbf{0.99} & 0.89 & 0.69 & \textbf{1.74} \\
\midrule
\multirow{2}{*}{Law} & Ours & \textbf{1.00} & \textbf{1.00} & \textbf{1.00} & \textbf{0.37} & \textbf{0.23} & \textbf{2.05} \\
 & BCE & \textbf{1.00} & \textbf{1.00} & \textbf{0.98} & 0.97 & 0.60 & 1.67 \\
\midrule
\multirow{2}{*}{Audit} & Ours & \textbf{1.00} & \textbf{0.99} & \textbf{0.99} & \textbf{2.04} & \textbf{0.79} & 51.64 \\
 & BCE & \textbf{1.00} & \textbf{0.99} & \textbf{0.98} & 3.01 & 1.25 & \textbf{52.54} \\
\midrule
\multirow{2}{*}{Heloc} & Ours & \textbf{1.00} & \textbf{0.99} & \textbf{0.99} & \textbf{0.85} & \textbf{0.23} & \textbf{37.50} \\
 & BCE & \textbf{1.00} & \textbf{0.97} & \textbf{0.99} & 1.91 & 0.54 & 34.50 \\
 \midrule
\multirow{2}{*}{Blobs} & Ours & \textbf{1.00} & \textbf{1.00} & \textbf{1.00} & \textbf{0.69} & \textbf{0.50} & \textbf{3.00} \\
 & CE & \textbf{1.00} & \textbf{1.00} & 0.93 & 0.82 & 0.60 & 2.85 \\
 \midrule
\multirow{2}{*}{Digits} & Ours & \textbf{1.00} & \textbf{1.00} & \textbf{1.00} & \textbf{8.27} & \textbf{1.33} & \textbf{44.42} \\
 & CE & \textbf{1.00} & \textbf{1.00} & \textbf{1.00} & 12.67 & 2.13 & 44.18 \\
 \midrule
\multirow{2}{*}{Wine} & Ours & \textbf{1.00} & \textbf{1.00} & \textbf{1.00} & \textbf{1.65} & \textbf{0.53} & \textbf{9.72} \\
 & CE & \textbf{1.00} & \textbf{1.00} & \textbf{0.99} & 3.87 & 1.29 & 9.29 \\
\bottomrule
\end{tabular}
\end{scriptsize}
\end{sc}
\end{center}
\end{table}

In this ablation study, we examined the influence of discriminative loss function selection on the effectiveness of our proposed method. While Binary Cross Entropy (BCE) and Cross Entropy (CE) losses are conventional choices for binary and multiclass problems, respectively, we compared them against our proposed discriminative loss function to understand their impacts on the results. The findings, detailed in Tab. \ref{tab:ablation_loss}, reveal a notable distinction in distance metrics. Our method, using the specialized loss function, demonstrated significantly better proximity to original observations compared to BCE and CE. This improvement is attributed to our loss function's design, which zeroes the classification component of the loss upon surpassing by $\epsilon$ a classification threshold. This allows for more rapid convergence to closer counterfactuals, while CE, by continually seeking points with higher classification confidence, tends to push counterfactuals further from the original samples. Consequently, this affects the final values in proximity metrics, underscoring the advantage of our approach in generating more proximate and plausible counterfactuals.

\subsection{Regularization Hyperparameter $\lambda$ Analysis}

\begin{table}[t]
\centering
\caption{Ablation Study on Regularization Hyperparameter $\lambda$.}
% We assess how different values of $\lambda$ influence the validity and probabilistic plausibility of counterfactual explanations. Our findings indicate that values around $100$ practically guarantee the fulfillment of the validity and probabilistic plausibility conditions.
\label{tab:ablation_alpha}

\begin{center}
\begin{sc}
\begin{scriptsize}
\begin{tabular}{l|l|rrrrrr}
\toprule
Dataset & $\lambda$ & Cov. & Val. & PP & L1 & L2 & LD \\
\midrule
\multirow{6}{*}{Moons}  & 1 & 1.00 & 0.46 & 0.78 & 0.43 & 0.34 & 1.61 \\
 & 2 & 1.00 & 0.95 & 0.92 & 0.43 & 0.34 & 1.63 \\
 & 5 & 1.00 & 0.99 & 0.98 & 0.43 & 0.34 & 1.66 \\
 & 10 & 1.00 & 0.99 & 1.00 & 0.44 & 0.35 & 1.70 \\
 & 100 & 1.00 & 1.00 & 1.00 & 0.45 & 0.36 & 1.70 \\
 & 1000 & 1.00 & 1.00 & 1.00 & 0.45 & 0.36 & 1.70 \\
 \midrule
\multirow{6}{*}{Law}  & 1 & 1.00 & 0.48 & 0.98 & 0.19 & 0.12 & 1.85 \\
 & 2 & 1.00 & 0.99 & 0.99 & 0.28 & 0.18 & 1.88 \\
 & 5 & 1.00 & 1.00 & 1.00 & 0.29 & 0.18 & 1.94 \\
 & 10 & 1.00 & 1.00 & 1.00 & 0.30 & 0.18 & 2.00 \\
 & 100 & 1.00 & 1.00 & 1.00 & 0.34 & 0.21 & 2.08 \\
 & 1000 & 1.00 & 1.00 & 1.00 & 0.38 & 0.22 & 2.09 \\
% \midrule
% \multirow{6}{*}{Audit}  & 1 & 1.0 & 0.38 & 0.96 & 1.61 & 0.59 & 42.69 \\
%  & 2 & 1.0 & 0.97 & 0.98 & 2.06 & 0.75 & 43.07 \\
%  & 5 & 1.0 & 0.99 & 0.97 & 2.21 & 0.83 & 42.58 \\
%  & 10 & 1.0 & 0.99 & 0.98 & 2.29 & 0.88 & 42.52 \\
%  & 100 & 1.0 & 0.99 & 0.99 & 2.51 & 0.99 & 42.57 \\
%  & 1000 & 1.0 & 0.99 & 1.0 & 2.23 & 0.85 & 42.62 \\ 
\bottomrule
\end{tabular}
\end{scriptsize}
\end{sc}
\end{center}
\end{table}

To evaluate the impact of the regularization hyperparameter $\lambda$ on the fulfillment of validity and probabilistic plausibility conditions, we conducted a focused hyperparameter sensitivity analysis. While $\lambda$ theoretically should extend to infinity, practical considerations necessitate setting a feasible value. Our objective is to identify an optimal $\lambda$ that not only guarantees condition fulfillment but also to understand its influence on other metrics. Experiments were carried out on the Moons and Law datasets, exploring $\lambda$ values within the set $\{1, 2, 5, 10, 100, 1000\}$. The results in Tab. \ref{tab:ablation_alpha} indicate that moderate values of $\lambda$, like $5$ or $10$, deliver satisfactory outcomes, while values around $100$ or more almost invariably guarantee the fulfillment of the conditions, leading us to adopt the value of $100$ for all preceding experiments. This experiment confirms the expected trade-off: higher strictness in counterfactual conditions leads to decreased proximity metrics, requiring larger deviations from the original data point.

\section{Conclusions}

In this work, we present \our{}, a novel method for generating counterfactual explanations that utilize normalizing flows as density estimators within an unconstrained optimization framework. This technique adeptly balances essential factors such as distance, validity, and probabilistic plausibility in the counterfactuals it produces. Notably, \our{} is computationally efficient and capable of handling large datasets, making it highly applicable in real-world scenarios. The method's flexible design allows for future enhancements, including other desirable counterfactual attributes like actionability or sparsity, and to generate plausible counterfactuals in label-scarce environments.

% \section*{Broader Impact}
% The development of a method for probabilistically plausible counterfactual explanations marks a significant advancement in the field of explainable artificial intelligence (XAI). This method holds the potential to enhance transparency and interpretability in complex machine learning models, particularly in high-stakes domains such as healthcare, finance, and legal systems. By providing explanations that are not only understandable but also grounded in realistic and probable scenarios, this method can facilitate better decision-making, foster trust among users, and ensure compliance with ethical and regulatory standards. Additionally, it can aid in identifying and mitigating biases in AI systems, thereby contributing to the development of more fair and equitable technologies. However, it's also crucial to be mindful of the potential misuse of such explanations in misleading or manipulating end-users, and hence, ethical guidelines and responsible use are paramount.

\section*{Acknowledgements}
Patryk Wielopolski, Oleksii Furman, and Maciej Zieba's work was supported by the National Science Centre (Poland) Grant No. 2021/43/B/ST6/02853, and Jerzy Stefanowski's work was supported by the National Science Centre (Poland) grant No. 2023/51/B/ST6/00545. Moreover, we gratefully acknowledge Polish high-performance computing infrastructure PLGrid (HPC Center: ACK Cyfronet AGH) for providing computer facilities and support within computational grant no. PLG/2023/016636.

\bibliography{main}

\begin{thebibliography}{34}
\providecommand{\natexlab}[1]{#1}
\providecommand{\url}[1]{\texttt{#1}}
\expandafter\ifx\csname urlstyle\endcsname\relax
  \providecommand{\doi}[1]{doi: #1}\else
  \providecommand{\doi}{doi: \begingroup \urlstyle{rm}\Url}\fi

\bibitem[Arrieta and Ser(2020)]{ArrietaS20}
A.~B. Arrieta and J.~D. Ser.
\newblock Plausible counterfactuals: Auditing deep learning classifiers with realistic adversarial examples.
\newblock In \emph{2020 International Joint Conference on Neural Networks, {IJCNN} 2020, Glasgow, United Kingdom, July 19-24, 2020}, pages 1--7. {IEEE}, 2020.

\bibitem[Artelt and Hammer(2019)]{ArteltH19}
A.~Artelt and B.~Hammer.
\newblock On the computation of counterfactual explanations - {A} survey.
\newblock \emph{CoRR}, abs/1911.07749, 2019.

\bibitem[Artelt and Hammer(2020)]{ArteltH20}
A.~Artelt and B.~Hammer.
\newblock Convex density constraints for computing plausible counterfactual explanations.
\newblock In \emph{Artificial Neural Networks and Machine Learning - {ICANN} 2020 - 29th International Conference on Artificial Neural Networks, Bratislava, Slovakia, September 15-18, 2020, Proceedings, Part {I}}, volume 12396 of \emph{Lecture Notes in Computer Science}, pages 353--365. Springer, 2020.

\bibitem[Augustin et~al.(2022)Augustin, Boreiko, Croce, and Hein]{AugustinBC022}
M.~Augustin, V.~Boreiko, F.~Croce, and M.~Hein.
\newblock Diffusion visual counterfactual explanations.
\newblock In \emph{Advances in Neural Information Processing Systems 35: Annual Conference on Neural Information Processing Systems 2022, NeurIPS 2022, New Orleans, LA, USA, November 28 - December 9, 2022}, 2022.

\bibitem[Dhurandhar et~al.(2018)Dhurandhar, Chen, Luss, Tu, Ting, Shanmugam, and Das]{DhurandharCLTTS18}
A.~Dhurandhar, P.~Chen, R.~Luss, C.~Tu, P.~Ting, K.~Shanmugam, and P.~Das.
\newblock Explanations based on the missing: Towards contrastive explanations with pertinent negatives.
\newblock In \emph{Advances in Neural Information Processing Systems 31: Annual Conference on Neural Information Processing Systems 2018, NeurIPS 2018, December 3-8, 2018, Montr{\'{e}}al, Canada}, pages 590--601, 2018.

\bibitem[Dinh et~al.(2015)Dinh, Krueger, and Bengio]{DinhKB14}
L.~Dinh, D.~Krueger, and Y.~Bengio.
\newblock {NICE:} non-linear independent components estimation.
\newblock In \emph{3rd International Conference on Learning Representations, {ICLR} 2015, San Diego, CA, USA, May 7-9, 2015, Workshop Track Proceedings}, 2015.

\bibitem[Dinh et~al.(2017)Dinh, Sohl{-}Dickstein, and Bengio]{DinhSB17}
L.~Dinh, J.~Sohl{-}Dickstein, and S.~Bengio.
\newblock Density estimation using real {NVP}.
\newblock In \emph{5th International Conference on Learning Representations, {ICLR} 2017, Toulon, France, April 24-26, 2017, Conference Track Proceedings}, 2017.

\bibitem[Dombrowski et~al.(2022)Dombrowski, Gerken, M{\"{u}}ller, and Kessel]{DomrowskiGMK22}
A.~Dombrowski, J.~E. Gerken, K.~M{\"{u}}ller, and P.~Kessel.
\newblock Diffeomorphic counterfactuals with generative models.
\newblock \emph{CoRR}, abs/2206.05075, 2022.
\newblock \doi{10.48550/ARXIV.2206.05075}.

\bibitem[Duong et~al.(2023)Duong, Li, and Xu]{DuongLX23}
T.~D. Duong, Q.~Li, and G.~Xu.
\newblock Ceflow: {A} robust and efficient counterfactual explanation framework for tabular data using normalizing flows.
\newblock In \emph{Advances in Knowledge Discovery and Data Mining - 27th Pacific-Asia Conference on Knowledge Discovery and Data Mining, {PAKDD} 2023, Osaka, Japan, May 25-28, 2023, Proceedings, Part {II}}, volume 13936 of \emph{Lecture Notes in Computer Science}, pages 133--144. Springer, 2023.

\bibitem[Guidotti(2022)]{Guidotti22}
R.~Guidotti.
\newblock Counterfactual explanations and how to find them: literature review and benchmarking.
\newblock \emph{Data Mining and Knowledge Discovery}, pages 1--55, 04 2022.

\bibitem[Jeanneret et~al.(2022)Jeanneret, Simon, and Jurie]{JeanneretSJ22}
G.~Jeanneret, L.~Simon, and F.~Jurie.
\newblock Diffusion models for counterfactual explanations.
\newblock In \emph{Computer Vision - {ACCV} 2022 - 16th Asian Conference on Computer Vision, Macao, China, December 4-8, 2022, Proceedings, Part {VII}}, volume 13847 of \emph{Lecture Notes in Computer Science}, pages 219--237. Springer, 2022.

\bibitem[Jiang et~al.(2023)Jiang, Lan, Leofante, Rago, and Toni]{JiangLLRT23}
J.~Jiang, J.~Lan, F.~Leofante, A.~Rago, and F.~Toni.
\newblock Provably robust and plausible counterfactual explanations for neural networks via robust optimisation.
\newblock \emph{CoRR}, abs/2309.12545, 2023.

\bibitem[Kanamori et~al.(2020)Kanamori, Takagi, Kobayashi, and Arimura]{KanamoriTKA20}
K.~Kanamori, T.~Takagi, K.~Kobayashi, and H.~Arimura.
\newblock {DACE:} distribution-aware counterfactual explanation by mixed-integer linear optimization.
\newblock In \emph{Proceedings of the Twenty-Ninth International Joint Conference on Artificial Intelligence, {IJCAI} 2020}, pages 2855--2862. ijcai.org, 2020.

\bibitem[Keane and Smyth(2020)]{KeaneS20}
M.~T. Keane and B.~Smyth.
\newblock Good counterfactuals and where to find them: {A} case-based technique for generating counterfactuals for explainable {AI} {(XAI)}.
\newblock In \emph{Case-Based Reasoning Research and Development - 28th International Conference, {ICCBR} 2020, Salamanca, Spain, June 8-12, 2020, Proceedings}, volume 12311 of \emph{Lecture Notes in Computer Science}, pages 163--178. Springer, 2020.

\bibitem[Looveren and Klaise(2021)]{LooverenK21}
A.~V. Looveren and J.~Klaise.
\newblock Interpretable counterfactual explanations guided by prototypes.
\newblock In \emph{Machine Learning and Knowledge Discovery in Databases. Research Track - European Conference, {ECML} {PKDD} 2021, Bilbao, Spain, September 13-17, 2021, Proceedings, Part {II}}, volume 12976 of \emph{Lecture Notes in Computer Science}, pages 650--665. Springer, 2021.

\bibitem[Mahajan et~al.(2019)Mahajan, Tan, and Sharma]{MahajanTS19}
D.~Mahajan, C.~Tan, and A.~Sharma.
\newblock Preserving causal constraints in counterfactual explanations for machine learning classifiers.
\newblock \emph{CoRR}, abs/1912.03277, 2019.

\bibitem[Mertes et~al.(2022)Mertes, Huber, Weitz, Heimerl, and André]{counterfactuals_medical}
S.~Mertes, T.~Huber, K.~Weitz, A.~Heimerl, and E.~André.
\newblock Ganterfactual—counterfactual explanations for medical non-experts using generative adversarial learning.
\newblock \emph{Frontiers in Artificial Intelligence}, 2022.

\bibitem[Mothilal et~al.(2020)Mothilal, Sharma, and Tan]{MothilalST20}
R.~K. Mothilal, A.~Sharma, and C.~Tan.
\newblock Explaining machine learning classifiers through diverse counterfactual explanations.
\newblock In \emph{FAT* '20: Conference on Fairness, Accountability, and Transparency, Barcelona, Spain, January 27-30, 2020}, pages 607--617. {ACM}, 2020.

\bibitem[Papamakarios et~al.(2017)Papamakarios, Murray, and Pavlakou]{PapamakariosMP17}
G.~Papamakarios, I.~Murray, and T.~Pavlakou.
\newblock Masked autoregressive flow for density estimation.
\newblock In \emph{Advances in Neural Information Processing Systems 30: Annual Conference on Neural Information Processing Systems 2017, December 4-9, 2017, Long Beach, CA, {USA}}, pages 2338--2347, 2017.

\bibitem[Paszke et~al.(2019)Paszke, Gross, Massa, Lerer, Bradbury, Chanan, Killeen, Lin, Gimelshein, Antiga, Desmaison, Kopf, Yang, DeVito, Raison, Tejani, Chilamkurthy, Steiner, Fang, Bai, and Chintala]{pytorch}
A.~Paszke, S.~Gross, F.~Massa, A.~Lerer, J.~Bradbury, G.~Chanan, T.~Killeen, Z.~Lin, N.~Gimelshein, L.~Antiga, A.~Desmaison, A.~Kopf, E.~Yang, Z.~DeVito, M.~Raison, A.~Tejani, S.~Chilamkurthy, B.~Steiner, L.~Fang, J.~Bai, and S.~Chintala.
\newblock Pytorch: An imperative style, high-performance deep learning library.
\newblock In \emph{Advances in Neural Information Processing Systems 32}, pages 8024--8035. Curran Associates, Inc., 2019.

\bibitem[Pearl et~al.(2016)Pearl, Glymour, and Jewell]{pearl2016causal}
J.~Pearl, M.~Glymour, and N.~Jewell.
\newblock \emph{Causal Inference in Statistics: A Primer}.
\newblock Wiley, 2016.
\newblock ISBN 9781119186847.

\bibitem[Popov et~al.(2020)Popov, Morozov, and Babenko]{PopovMB20}
S.~Popov, S.~Morozov, and A.~Babenko.
\newblock Neural oblivious decision ensembles for deep learning on tabular data.
\newblock In \emph{8th International Conference on Learning Representations, {ICLR} 2020, Addis Ababa, Ethiopia, April 26-30, 2020}. OpenReview.net, 2020.

\bibitem[Poyiadzi et~al.(2020)Poyiadzi, Sokol, Santos{-}Rodr{\'{\i}}guez, Bie, and Flach]{PoyiadziSSBF20}
R.~Poyiadzi, K.~Sokol, R.~Santos{-}Rodr{\'{\i}}guez, T.~D. Bie, and P.~A. Flach.
\newblock {FACE:} feasible and actionable counterfactual explanations.
\newblock In \emph{{AIES} '20: {AAAI/ACM} Conference on AI, Ethics, and Society, New York, NY, USA, February 7-8, 2020}, pages 344--350. {ACM}, 2020.

\bibitem[Rezende and Mohamed(2015)]{rezende2015variational}
D.~Rezende and S.~Mohamed.
\newblock Variational inference with normalizing flows.
\newblock In \emph{International Conference on Machine Learning}, pages 1530--1538. PMLR, 2015.

\bibitem[Russell(2019)]{Russell19}
C.~Russell.
\newblock Efficient search for diverse coherent explanations.
\newblock In danah boyd and J.~H. Morgenstern, editors, \emph{Proceedings of the Conference on Fairness, Accountability, and Transparency, FAT* 2019, Atlanta, GA, USA, January 29-31, 2019}, pages 20--28. {ACM}, 2019.

\bibitem[Shakhnarovich et~al.(2008)Shakhnarovich, Darrell, and Indyk]{ShakhnarovichDI08}
G.~Shakhnarovich, T.~Darrell, and P.~Indyk.
\newblock Nearest-neighbor methods in learning and vision.
\newblock \emph{{IEEE} Trans. Neural Networks}, 19\penalty0 (2):\penalty0 377, 2008.

\bibitem[Smyth and Keane(2022)]{SmythK22}
B.~Smyth and M.~T. Keane.
\newblock A few good counterfactuals: Generating interpretable, plausible and diverse counterfactual explanations.
\newblock In \emph{Case-Based Reasoning Research and Development - 30th International Conference, {ICCBR} 2022, Nancy, France, September 12-15, 2022, Proceedings}, volume 13405 of \emph{Lecture Notes in Computer Science}, pages 18--32. Springer, 2022.

\bibitem[Van~Rossum and Drake~Jr(1995)]{van1995python}
G.~Van~Rossum and F.~L. Drake~Jr.
\newblock \emph{Python reference manual}.
\newblock Centrum voor Wiskunde en Informatica Amsterdam, 1995.

\bibitem[Vercheval and Pizurica(2021)]{VerchevalP21}
N.~Vercheval and A.~Pizurica.
\newblock Hierarchical variational autoencoders for visual counterfactuals.
\newblock In \emph{2021 {IEEE} International Conference on Image Processing, {ICIP} 2021, Anchorage, AK, USA, September 19-22, 2021}, pages 2513--2517. {IEEE}, 2021.

\bibitem[Verma et~al.(2020)Verma, Boonsanong, Hoang, Hines, Dickerson, and Shah]{Verma2020}
S.~Verma, V.~Boonsanong, M.~Hoang, K.~E. Hines, J.~P. Dickerson, and C.~Shah.
\newblock Counterfactual explanations and algorithmic recourses for machine learning: A review.
\newblock \emph{arXiv preprint arXiv:2010.10596}, 2020.

\bibitem[Wachter et~al.(2017)Wachter, Mittelstadt, and Russell]{WachterMR17}
S.~Wachter, B.~D. Mittelstadt, and C.~Russell.
\newblock Counterfactual explanations without opening the black box: Automated decisions and the {GDPR}.
\newblock \emph{CoRR}, abs/1711.00399, 2017.

\bibitem[Wellawatte et~al.(2022)Wellawatte, Seshadri, and White]{counterfactuals_chemistry}
G.~P. Wellawatte, A.~Seshadri, and A.~D. White.
\newblock Model agnostic generation of counterfactual explanations for molecules.
\newblock \emph{Chem. Sci.}, 2022.

\bibitem[Wielopolski et~al.(2024)Wielopolski, Furman, Stefanowski, and Zieba]{abs-2405-17640}
P.~Wielopolski, O.~Furman, J.~Stefanowski, and M.~Zieba.
\newblock Probabilistically plausible counterfactual explanations with normalizing flows.
\newblock \emph{CoRR}, abs/2405.17640, 2024.
\newblock \doi{10.48550/ARXIV.2405.17640}.

\bibitem[Wightman(1998)]{Wightman98}
L.~F. Wightman.
\newblock Lsac national longitudinal bar passage study. lsac research report series.
\newblock Technical report, Law School Admission Council, Newtown, PA., 1998.

\end{thebibliography}

%%%%%%%%%%%%%%%%%%%%%%%%%%%%%%%%%%%%%%%%%%%%%%%%%%%%%%%%%%%%%%%%%%%%%%%%%%%%%%%
%%%%%%%%%%%%%%%%%%%%%%%%%%%%%%%%%%%%%%%%%%%%%%%%%%%%%%%%%%%%%%%%%%%%%%%%%%%%%%%
% APPENDIX
%%%%%%%%%%%%%%%%%%%%%%%%%%%%%%%%%%%%%%%%%%%%%%%%%%%%%%%%%%%%%%%%%%%%%%%%%%%%%%%
%%%%%%%%%%%%%%%%%%%%%%%%%%%%%%%%%%%%%%%%%%%%%%%%%%%%%%%%%%%%%%%%%%%%%%%%%%%%%%%

\appendix
\onecolumn

\section{Additional Results}
\label{apx:additional_results}
This section supplements the main manuscript's experimental results with a more comprehensive analysis. We include means and standard deviations from five-fold cross-validation for greater statistical rigor and broaden the comparison with additional datasets and metrics.
Tab. \ref{tab:ours_vs_all_cv_lr} compares the effectiveness and nuances of different approaches in generating counterfactual explanations for the Logistic Regression model. In Tab. \ref{tab:ours_vs_all_cv_mlp}, we delve into Multilayer Perceptron, presenting a similar analysis that highlights the unique aspects and performance metrics relevant to this model. Moving forward, we examine the performance of counterfactual methods for deep ensembles of oblivious differentiable decision trees. Results for NODE classifier are presented in Tab. \ref{tab:ours_vs_all_cv_node}. 
Notably, Artelt’s method is incompatible with non-linear classifiers such as Multilayer Perceptrons (MLP) or NODE, precluding the acquisition of performance data for these models using this approach. Furthermore, the application of CEGP to the NODE classifier was prohibitively time-consuming, which prevented the generation of results for this method as well.
In Tab. \ref{tab:ablation_loss_cv}, we provide extended results for the Ablation Study on Loss Function with additional metrics: LOF and Isolation Forest.

\begin{landscape}
\begin{table*}[t]
\centering
\caption{Detailed Comparative Results of Probabilistically Plausible Counterfactual Explanation Methods for \textbf{Logistic Regression} classifier. We present a comprehensive comparison of our method with other established reference methods across various datasets. The results presented include the mean and standard deviation obtained from a five-fold cross-validation.}
\label{tab:ours_vs_all_cv_lr}

\begin{center}
\begin{sc}
\begin{scriptsize}
\begin{tabular}{l|l|rrrrrrrrr}
\toprule
Dataset & Method & Coverage $\uparrow$ & Validity $\uparrow$ & Prob. Plaus. $\uparrow$ & LOF & IsoForest & Log Dens. $\uparrow$ & L1 $\downarrow$ & L2 $\downarrow$ & Time $\downarrow$ \\
\midrule
\multirow{6}{*}{Moons} & CBCE & \textbf{1.00$\pm$0.00} & \textbf{1.00$\pm$0.00} & 0.10$\pm$0.23 & 1.06$\pm$0.02 & 0.03$\pm$0.00 & -5.81$\pm$3.74 & 0.62$\pm$0.07 & 0.48$\pm$0.05 & \textbf{0.07$\pm$0.01} s\\
 & CEGP & \textbf{1.00$\pm$0.00} & \textbf{1.00$\pm$0.00} & 0.09$\pm$0.04 & 1.36$\pm$0.03 & 0.00$\pm$0.00 & -6.66$\pm$0.82 & 0.36$\pm$0.02 & \textbf{0.28$\pm$0.01} & 904.11$\pm$11.12 s\\
 & CEM & \textbf{1.00$\pm$0.00} & \textbf{1.00$\pm$0.00} & 0.14$\pm$0.03 & 2.03$\pm$0.12 & -0.07$\pm$0.01 & -10.09$\pm$6.62 & 0.55$\pm$0.03 & 0.50$\pm$0.02 & 211.56$\pm$1.50 s\\
 & WACH & 0.98$\pm$0.03 & \textbf{1.00$\pm$0.01} & 0.11$\pm$0.05 & 1.55$\pm$0.08 & -0.01$\pm$0.00 & -6.34$\pm$2.41 & 0.49$\pm$0.02 & 0.36$\pm$0.01 & 198.29$\pm$3.66 s\\
 & ARTELT & \textbf{1.00$\pm$0.00} & \textbf{1.00$\pm$0.00} & 0.08$\pm$0.03 & 1.53$\pm$0.09 & -0.03$\pm$0.01 & -8.74$\pm$3.57 & \textbf{0.32$\pm$0.02} & 0.32$\pm$0.02 & 4.15$\pm$0.69 s\\
 & \our{} & \textbf{1.00$\pm$0.00} & \textbf{1.00$\pm$0.00} & \textbf{1.00$\pm$0.00} & 1.01$\pm$0.02 & 0.04$\pm$0.01 & \textbf{1.69$\pm$0.07} & 0.45$\pm$0.01 & 0.36$\pm$0.01 & 1.85$\pm$0.01 s\\
\midrule
\multirow{6}{*}{Law} & CBCE & \textbf{1.00$\pm$0.00} & \textbf{1.00$\pm$0.00} & 0.49$\pm$0.35 & 1.05$\pm$0.02 & 0.04$\pm$0.02 & 1.28$\pm$0.41 & 0.61$\pm$0.03 & 0.40$\pm$0.02 & \textbf{0.23$\pm$0.00} s\\
 & CEGP & \textbf{1.00$\pm$0.00} & \textbf{1.00$\pm$0.00} & 0.49$\pm$0.04 & 1.07$\pm$0.00 & 0.04$\pm$0.00 & 1.08$\pm$0.04 & 0.23$\pm$0.01 & \textbf{0.18$\pm$0.01} & 1973.76$\pm$11.09 s\\
 & CEM & \textbf{1.00$\pm$0.00} & \textbf{1.00$\pm$0.00} & 0.26$\pm$0.01 & 1.26$\pm$0.00 & -0.02$\pm$0.00 & -0.56$\pm$0.08 & 0.33$\pm$0.01 & 0.31$\pm$0.01 & 368.10$\pm$51.57 s\\
 & WACH & \textbf{1.00$\pm$0.00} & \textbf{1.00$\pm$0.00} & 0.39$\pm$0.03 & 1.30$\pm$0.02 & -0.01$\pm$0.00 & -0.29$\pm$0.13 & 0.45$\pm$0.01 & 0.35$\pm$0.01 & 359.00$\pm$41.37 s\\
 & ARTELT & \textbf{1.00$\pm$0.00} & \textbf{1.00$\pm$0.00} & 0.40$\pm$0.02 & 1.12$\pm$0.01 & 0.02$\pm$0.00 & 0.54$\pm$0.08 & \textbf{0.20$\pm$0.01} & 0.20$\pm$0.01 & 4.02$\pm$0.42 s\\
 & \our{} & \textbf{1.00$\pm$0.00} & \textbf{1.00$\pm$0.00} & \textbf{1.00$\pm$0.00} & 1.03$\pm$0.00 & 0.07$\pm$0.00 & \textbf{2.05$\pm$0.02} & 0.37$\pm$0.01 & 0.23$\pm$0.01 & 2.42$\pm$0.10 s\\
\midrule
\multirow{6}{*}{Audit} & CBCE & \textbf{1.00$\pm$0.00} & \textbf{1.00$\pm$0.00} & 0.79$\pm$0.28 & 11.70$\pm$20.10 & 0.14$\pm$0.00 & \textbf{54.97$\pm$3.89} & 2.55$\pm$0.18 & 1.24$\pm$0.10 & \textbf{0.04$\pm$0.01} s\\
 & CEGP & 0.97$\pm$0.02 & \textbf{1.00$\pm$0.00} & 0.02$\pm$0.03 & 6.08$\cdot\text{10}^7\pm$5.58$\cdot\text{10}^7$ & 0.02$\pm$0.02 & 8.09$\pm$13.27 & 1.56$\pm$0.13 & \textbf{0.57$\pm$0.06} & 561.04$\pm$13.04 s\\
 & CEM & 0.52$\pm$0.03 & \textbf{1.00$\pm$0.00} & 0.00$\pm$0.01 & 8.28$\cdot\text{10}^6\pm$1.85$\cdot\text{10}^7$ & -0.04$\pm$0.03 & 20.84$\pm$10.32 & 1.20$\pm$0.06 & 0.37$\pm$0.02 & 105.92$\pm$13.20 s\\
 & WACH & \textbf{0.99$\pm$0.01} & \textbf{1.00$\pm$0.00} & 0.02$\pm$0.02 & 1.42$\cdot\text{10}^8\pm$2.99$\cdot\text{10}^7$ & 0.06$\pm$0.01 & -40.34$\pm$34.83 & 1.78$\pm$0.08 & 0.80$\pm$0.05 & 101.27$\pm$10.95 s\\
 & ARTELT & 0.60$\pm$0.22 & 0.97$\pm$0.05 & 0.00$\pm$0.01 & 4.09$\cdot\text{10}^8\pm$5.89$\cdot\text{10}^8$ & 0.10$\pm$0.02 & -3585.76$\pm$7834.29 & \textbf{0.90$\pm$0.14} & 0.88$\pm$0.10 & 43.84$\pm$31.22 s\\
 & \our{} & \textbf{1.00$\pm$0.00} & \textbf{0.99$\pm$0.01} & \textbf{0.99$\pm$0.01} & 4.25$\cdot\text{10}^7\pm$9.32$\cdot\text{10}^7$ & 0.08$\pm$0.01 & 51.64$\pm$4.53 & 2.04$\pm$0.15 & 0.79$\pm$0.12 & 7.01$\pm$1.08 s\\
\midrule
\multirow{6}{*}{Heloc} & CBCE & \textbf{1.00$\pm$0.00} & \textbf{1.00$\pm$0.00} & 0.54$\pm$0.01 & 1.10$\pm$0.08 & 0.07$\pm$0.03 & 28.01$\pm$3.31 & 2.84$\pm$0.39 & 0.82$\pm$0.11 & \textbf{5.71$\pm$0.41} s\\
 & CEGP & \textbf{1.00$\pm$0.00} & \textbf{1.00$\pm$0.00} & 0.29$\pm$0.03 & 3.50$\cdot\text{10}^7\pm$7.28$\cdot\text{10}^6$ & 0.04$\pm$0.00 & 24.75$\pm$0.52 & \textbf{0.26$\pm$0.03} & \textbf{0.10$\pm$0.01} & 9654.60$\pm$81.96 s\\
 & CEM & \textbf{1.00$\pm$0.00} & \textbf{1.00$\pm$0.00} & 0.07$\pm$0.01 & 2.50$\cdot\text{10}^8\pm$4.00$\cdot\text{10}^7$ & 0.02$\pm$0.01 & 12.37$\pm$2.74 & 0.35$\pm$0.05 & 0.20$\pm$0.02 & 1639.16$\pm$9.35 s\\
 & WACH & \textbf{1.00$\pm$0.00} & \textbf{1.00$\pm$0.00} & 0.00$\pm$0.00 & 2.65$\cdot\text{10}^8\pm$3.04$\cdot\text{10}^7$ & 0.03$\pm$0.01 & -15.09$\pm$5.86 & 0.74$\pm$0.06 & 0.37$\pm$0.01 & 1600.28$\pm$17.36 s\\
 & ARTELT & - & - & - & - & - & - & - & - & - \\
 & \our{} & \textbf{1.00$\pm$0.00} & \textbf{1.00$\pm$0.00} & \textbf{1.00$\pm$0.00} & 6.47$\cdot\text{10}^7$$\pm$2.16$\cdot\text{10}^7$ & 0.07$\pm$0.00 & \textbf{32.42$\pm$0.34} & 0.90$\pm$0.03 & 0.23$\pm$0.01 & 12.44$\pm$2.36 s\\
\midrule
\multirow{6}{*}{Blobs} & CBCE & \textbf{1.00$\pm$0.00} & \textbf{1.00$\pm$0.00} & 0.27$\pm$0.15 & 1.02$\pm$0.03 & 0.03$\pm$0.02 & -35.52$\pm$15.68 & 0.95$\pm$0.01 & 0.72$\pm$0.01 & \textbf{0.13$\pm$0.00} s\\
 & CEGP & \textbf{1.00$\pm$0.00} & \textbf{1.00$\pm$0.00} & 0.00$\pm$0.00 & 2.43$\pm$0.12 & -0.07$\pm$0.01 & -9.08$\pm$2.29 & \textbf{0.30$\pm$0.01} & \textbf{0.25$\pm$0.01} & 1295.36$\pm$30.32 s\\
 & CEM & 0.96$\pm$0.02 & \textbf{1.00$\pm$0.00} & 0.00$\pm$0.00 & 3.51$\pm$0.09 & -0.12$\pm$0.00 & -14.95$\pm$2.91 & 0.46$\pm$0.04 & 0.45$\pm$0.03 & 512.56$\pm$5.56 s\\
 & WACH & \textbf{1.00$\pm$0.00} & \textbf{1.00$\pm$0.00} & 0.04$\pm$0.02 & 2.24$\pm$0.19 & -0.06$\pm$0.01 & -9.52$\pm$1.80 & 0.51$\pm$0.02 & 0.38$\pm$0.02 & 441.59$\pm$15.39 s\\
 & ARTELT & \textbf{1.00$\pm$0.00} & \textbf{1.00$\pm$0.00} & 0.00$\pm$0.00 & 2.11$\pm$0.16 & -0.07$\pm$0.01 & -3.51$\pm$0.33 & 0.39$\pm$0.01 & 0.33$\pm$0.01 & 6.62$\pm$0.09 s\\
 & \our{} & \textbf{1.00$\pm$0.00} & \textbf{1.00$\pm$0.00} & \textbf{1.00$\pm$0.00} & 1.01$\pm$0.01 & 0.04$\pm$0.01 & \textbf{3.00$\pm$0.11} & 0.69$\pm$0.05 & 0.50$\pm$0.04 & 3.22$\pm$0.84 s\\
\midrule
\multirow{6}{*}{Digits} & CBCE & \textbf{1.00$\pm$0.00} & \textbf{1.00$\pm$0.00} & 0.18$\pm$0.04 & 1.02$\pm$0.06 & 0.04$\pm$0.00 & 23.72$\pm$4.73 & 16.28$\pm$0.62 & 3.09$\pm$0.02 & \textbf{0.51$\pm$0.16} s\\
 & CEGP & \textbf{1.00$\pm$0.00} & \textbf{1.00$\pm$0.00} & 0.11$\pm$0.03 & 1.09$\pm$0.01 & 0.01$\pm$0.00 & -0.39$\pm$4.80 & 2.53$\pm$0.11 & \textbf{0.63$\pm$0.02} & 1945.67$\pm$22.30 s\\
 & CEM & \textbf{1.00$\pm$0.00} & 0.98$\pm$0.04 & 0.01$\pm$0.00 & 1.23$\pm$0.01 & -0.03$\pm$0.01 & -86.77$\pm$17.24 & 5.28$\pm$4.93 & 1.38$\pm$0.75 & 852.05$\pm$27.68 s\\
 & WACH & \textbf{1.00$\pm$0.00} & \textbf{1.00$\pm$0.00} & 0.08$\pm$0.03 & 1.20$\pm$0.01 & 0.00$\pm$0.01 & -34.97$\pm$7.01 & \textbf{2.47$\pm$0.05} & 1.20$\pm$0.02 & 651.00$\pm$7.97 s\\
 & ARTELT & 0.80$\pm$0.12 & 0.93$\pm$0.02 & 0.04$\pm$0.03 & 1.69$\pm$0.05 & 0.01$\pm$0.01 & -54.72$\pm$11.81 & 3.30$\pm$0.13 & 2.43$\pm$0.14 & 238.28$\pm$33.28 s\\
 & \our{} & \textbf{1.00$\pm$0.00} & \textbf{1.00$\pm$0.00} & \textbf{1.00$\pm$0.00} & 1.12$\pm$0.01 & 0.03$\pm$0.01 & \textbf{44.42$\pm$1.87} & 8.27$\pm$0.24 & 1.33$\pm$0.04 & 8.68$\pm$3.65 s\\
\midrule
\multirow{6}{*}{Wine} & CBCE & \textbf{1.00$\pm$0.00} & \textbf{1.00$\pm$0.00} & 0.37$\pm$0.05 & 1.06$\pm$0.05 & 0.05$\pm$0.02 & 2.13$\pm$1.28 & 3.38$\pm$0.14 & 1.12$\pm$0.05 & \textbf{0.01$\pm$0.00} s\\
 & CEGP & \textbf{1.00$\pm$0.00} & \textbf{1.00$\pm$0.00} & 0.01$\pm$0.01 & 1.08$\pm$0.03 & 0.05$\pm$0.01 & -0.15$\pm$1.54 & 0.82$\pm$0.08 & \textbf{0.32$\pm$0.03} & 191.09$\pm$4.00 s\\
 & CEM & \textbf{1.00$\pm$0.00} & \textbf{1.00$\pm$0.00} & 0.00$\pm$0.00 & 1.35$\pm$0.03 & -0.02$\pm$0.01 & -12.94$\pm$1.94 & 1.20$\pm$0.09 & 0.63$\pm$0.05 & 81.33$\pm$1.88 s\\
 & WACH & \textbf{1.00$\pm$0.00} & \textbf{1.00$\pm$0.00} & 0.01$\pm$0.01 & 1.27$\pm$0.04 & 0.00$\pm$0.01 & -9.41$\pm$1.95 & 1.57$\pm$0.10 & 0.78$\pm$0.05 & 50.74$\pm$5.63 s\\
 & ARTELT & \textbf{1.00$\pm$0.00} & 0.97$\pm$0.04 & 0.01$\pm$0.02 & 1.33$\pm$0.03 & 0.02$\pm$0.00 & -11.73$\pm$2.36 & \textbf{0.65$\pm$0.05} & 0.96$\pm$0.08 s\\
 & \our{} & \textbf{1.00$\pm$0.00} & \textbf{1.00$\pm$0.00} & \textbf{1.00$\pm$0.00} & 1.01$\pm$0.01 & 0.09$\pm$0.01 & \textbf{9.72$\pm$0.62} & 1.65$\pm$0.09 & 0.53$\pm$0.04 & 2.03$\pm$0.47 s\\
\bottomrule
\end{tabular}
\end{scriptsize}
\end{sc}
\end{center}
\end{table*}
\end{landscape}

\begin{landscape}
\begin{table*}[t]
\centering
\caption{Detailed Comparative Results of Probabilistically Plausible Counterfactual Explanation Methods for \textbf{Multilayer Perceptron} classifier. We provide an in-depth comparison between our approach and other well-established methods, utilizing a range of datasets. This analysis includes both the average values and the standard deviations derived from five-fold cross-validation.}
\label{tab:ours_vs_all_cv_mlp}

\begin{center}
\begin{sc}
\begin{scriptsize}
\begin{tabular}{l|l|rrrrrrrrr}
\toprule
Dataset & Method & Coverage $\uparrow$ & Validity $\uparrow$ & Prob. Plaus. $\uparrow$ & LOF & IsoForest & Log Dens. $\uparrow$ & L1 $\downarrow$ & L2 $\downarrow$ & Time $\downarrow$ \\
\midrule
\multirow{5}{*}{Moons} & CBCE & \textbf{1.00$\pm$0.00} & 0.84$\pm$0.24 & 0.58$\pm$0.16 & 1.03$\pm$0.03 & 0.02$\pm$0.01 & 1.23$\pm$0.27 & 0.71$\pm$0.16 & 0.53$\pm$0.11 & \textbf{0.08$\pm$0.00} s\\
 & CEGP & 0.97$\pm$0.03 & 0.33$\pm$0.13 & 0.06$\pm$0.07 & 1.39$\pm$0.10 & 0.00$\pm$0.01 & -3.67$\pm$1.16 & 0.29$\pm$0.07 & \textbf{0.23$\pm$0.05} & 1562.34$\pm$119.65 s\\
 & CEM & 0.99$\pm$0.02 & 0.45$\pm$0.15 & 0.03$\pm$0.04 & 2.29$\pm$0.16 & -0.08$\pm$0.01 & -11.92$\pm$6.48 & 0.49$\pm$0.04 & 0.48$\pm$0.04 & 784.34$\pm$33.83 s\\
 & WACH & 1.00$\pm$0.01 & 0.56$\pm$0.06 & 0.04$\pm$0.05 & 1.52$\pm$0.12 & -0.01$\pm$0.01 & -3.71$\pm$2.54 & \textbf{0.28$\pm$0.09} & 0.24$\pm$0.08 & 1452.01$\pm$99.08 s\\
 & ARTELT & - & - & - & - & - & - & - & - & - s\\
 & \our{} & \textbf{1.00$\pm$0.00} & \textbf{0.98$\pm$0.01} & \textbf{1.00$\pm$0.00} & 0.99$\pm$0.02 & 0.03$\pm$0.01 & \textbf{1.62$\pm$0.04} & 0.44$\pm$0.05 & 0.34$\pm$0.04 & 20.44$\pm$1.75 s\\
\midrule
\multirow{5}{*}{Law} & CBCE & \textbf{1.00$\pm$0.00} & 0.79$\pm$0.11 & 0.36$\pm$0.40 & 1.05$\pm$0.02 & 0.03$\pm$0.03 & 1.18$\pm$0.45 & 0.67$\pm$0.08 & 0.44$\pm$0.05 & \textbf{0.28$\pm$0.02} s\\
 & CEGP & 0.93$\pm$0.02 & 0.28$\pm$0.01 & 0.53$\pm$0.05 & 1.07$\pm$0.00 & 0.04$\pm$0.00 & 1.23$\pm$0.16 & \textbf{0.24$\pm$0.02} & \textbf{0.17$\pm$0.01} & 2986.44$\pm$60.71 s\\
& CEM & \textbf{1.00$\pm$0.00} & 0.61$\pm$0.04 & 0.27$\pm$0.01 & 1.26$\pm$0.01 & -0.02$\pm$0.00 & -0.22$\pm$0.11 & 0.29$\pm$0.01 & 0.28$\pm$0.01 & 1413.03$\pm$97.23 s\\
 & WACH & \textbf{1.00$\pm$0.00} & 0.74$\pm$0.04 & 0.42$\pm$0.03 & 1.14$\pm$0.01 & 0.01$\pm$0.00 & 0.58$\pm$0.09 & 0.38$\pm$0.01 & 0.29$\pm$0.01 & 2459.88$\pm$78.30 s\\
 & ARTELT & - & - & - & - & - & - & - & - & - s\\
 & \our{} & \textbf{1.00$\pm$0.00} & \textbf{0.95$\pm$0.01} & \textbf{1.00$\pm$0.00} & 1.03$\pm$0.00 & 0.07$\pm$0.00 & \textbf{2.04$\pm$0.02} & 0.40$\pm$0.01 & 0.24$\pm$0.01 & 20.63$\pm$1.08 s\\
\midrule
\multirow{5}{*}{Audit} & CBCE & \textbf{1.00$\pm$0.00} & 0.93$\pm$0.07 & 0.61$\pm$0.21 & 11.10$\pm$13.30 & 0.14$\pm$0.00 & 55.02$\pm$3.49 & 2.77$\pm$0.25 & 1.34$\pm$0.14 & \textbf{0.16$\pm$0.01} s\\
 & CEGP & \textbf{0.55$\pm$0.02} & 0.48$\pm$0.02 & 0.02$\pm$0.03 & 1.45$\cdot\text{10}^2\pm$3.86$\cdot\text{10}^1$ & 0.00$\pm$0.03 & -85.64$\pm$41.77 & 1.42$\pm$0.12 & \textbf{0.57$\pm$0.03} & 1426.33$\pm$55.75 s\\
 & CEM & 0.53$\pm$0.02 & 0.54$\pm$0.05 & 0.01$\pm$0.01 & 6.52$\cdot\text{10}^7\pm$8.76$\cdot\text{10}^7$ & 0.04$\pm$0.03 & -132.43$\pm$98.13 & \textbf{1.19$\pm$0.14} & 0.61$\pm$0.04 & 441.85$\pm$7.56 s\\
 & WACH & 0.49$\pm$0.02 & 0.78$\pm$0.06 & 0.00$\pm$0.00 & 1.24$\cdot\text{10}^2\pm$2.75$\cdot\text{10}^1$ & -0.03$\pm$0.03 & -155.20$\pm$74.93 & 1.49$\pm$0.07 & 0.65$\pm$0.01 & 601.77$\pm$23.34 s\\
 & ARTELT & - & - & - & - & - & - & - & - & - s\\
 & \our{} & \textbf{1.00$\pm$0.00} & \textbf{0.99$\pm$0.02} & \textbf{0.99$\pm$0.01} & 1.44$\cdot\text{10}^7\pm$1.88$\cdot\text{10}^7$ & 0.08$\pm$0.01 & 51.72$\pm$4.59 & 2.14$\pm$0.11 & 0.83$\pm$0.10 & 46.60$\pm$3.82 s\\
\midrule
\multirow{5}{*}{Heloc} & CBCE & \textbf{1.00$\pm$0.00} & \textbf{0.94$\pm$0.01} & 0.54$\pm$0.03 & 1.09$\pm$0.08 & 0.08$\pm$0.03 & 28.85$\pm$3.95 & 2.87$\pm$0.40 & 0.82$\pm$0.11 & \textbf{6.47$\pm$0.69} s\\
 & CEGP & 0.94$\pm$0.01 & 0.63$\pm$0.01 & 0.05$\pm$0.01 & 4.15$\cdot\text{10}^8\pm$1.97$\cdot\text{10}^8$ & 0.01$\pm$0.00 & -3.28$\pm$4.79 & 1.25$\pm$0.11 & 0.43$\pm$0.03 & 31309.33$\pm$3342.40 s\\
 & CEM & \textbf{1.00$\pm$0.00} & 0.86$\pm$0.01 & 0.01$\pm$0.00 & 7.71$\cdot\text{10}^8\pm$1.41$\cdot\text{10}^8$ & -0.01$\pm$0.01 & -89.39$\pm$56.89 & 1.32$\pm$0.22 & 0.58$\pm$0.07 & 6938.45$\pm$79.08 s\\
 & WACH & 0.99$\pm$0.00 & 0.81$\pm$0.03 & 0.00$\pm$0.00 & 1.34$\cdot\text{10}^8\pm$4.44$\cdot\text{10}^7$ & -0.06$\pm$0.01 & -161.68$\pm$104.43 & 3.11$\pm$0.24 & 0.90$\pm$0.04 & 23392.40$\pm$3677.14 s\\
 & ARTELT & - & - & - & - & - & - & - & - & - s\\
 & \our{} & \textbf{1.00$\pm$0.00} & 0.92$\pm$0.03 & \textbf{1.00$\pm$0.00} & 1.42$\cdot\text{10}^8\pm$3.90$\cdot\text{10}^7$ & 0.07$\pm$0.00 & \textbf{32.07$\pm$0.63} & \textbf{1.18$\pm$0.11} & \textbf{0.31$\pm$0.03} & 25.32$\pm$6.41 s\\
\midrule
\multirow{5}{*}{Blobs} & CBCE & \textbf{1.00$\pm$0.00} & \textbf{1.00$\pm$0.00} & 0.27$\pm$0.15 & 1.02$\pm$0.03 & 0.03$\pm$0.02 & -35.45$\pm$15.48 & 0.95$\pm$0.01 & 0.72$\pm$0.01 & \textbf{0.16$\pm$0.00} s\\
 & CEGP & 0.98$\pm$0.02 & 0.47$\pm$0.04 & 0.00$\pm$0.00 & 2.48$\pm$0.16 & -0.07$\pm$0.00 & -9.54$\pm$3.30 & 0.35$\pm$0.01 & \textbf{0.29$\pm$0.01} & 3047.79$\pm$38.09 s\\
 & CEM & 0.97$\pm$0.04 & 0.71$\pm$0.07 & 0.00$\pm$0.00 & 3.36$\pm$0.12 & -0.11$\pm$0.00 & -18.58$\pm$10.39 & 0.46$\pm$0.02 & 0.45$\pm$0.01 & 1046.35$\pm$14.92 s\\
 & WACH & 0.99$\pm$0.01 & 0.57$\pm$0.04 & 0.00$\pm$0.00 & 2.67$\pm$0.15 & -0.07$\pm$0.01 & -10.72$\pm$3.68 & \textbf{0.34$\pm$0.01} & \textbf{0.29$\pm$0.01} & 1731.10$\pm$62.81 s\\
 & ARTELT & - & - & - & - & - & - & - & - & - s\\
 & \our{} & \textbf{1.00$\pm$0.00} & \textbf{1.00$\pm$0.00} & \textbf{1.00$\pm$0.00} & 1.03$\pm$0.02 & 0.04$\pm$0.00 & \textbf{2.91$\pm$0.04} & 0.65$\pm$0.01 & 0.47$\pm$0.01 & 19.55$\pm$0.30 s\\
\midrule
\multirow{5}{*}{Digits} & CBCE & \textbf{1.00$\pm$0.00} & \textbf{1.00$\pm$0.00} & 0.18$\pm$0.04 & 1.02$\pm$0.06 & 0.04$\pm$0.00 & 23.66$\pm$3.76 & 16.29$\pm$0.61 & 3.09$\pm$0.02 & \textbf{0.54$\pm$0.16} s\\
 & CEGP & 0.95$\pm$0.01 & 0.46$\pm$0.03 & 0.02$\pm$0.01 & 1.24$\pm$0.00 & -0.03$\pm$0.00 & -138.62$\pm$14.93 & 6.39$\pm$0.16 & \textbf{1.42$\pm$0.03} & 2523.28$\pm$54.03 s\\
 & CEM & \textbf{1.00$\pm$0.00} & 0.42$\pm$0.01 & 0.01$\pm$0.01 & 1.44$\pm$0.01 & -0.06$\pm$0.01 & -481.57$\pm$68.49 & \textbf{6.34$\pm$0.55} & 1.76$\pm$0.09 & 1260.54$\pm$51.03 s\\
 & WACH & \textbf{1.00$\pm$0.00} & 0.72$\pm$0.03 & 0.00$\pm$0.00 & 1.50$\pm$0.03 & -0.07$\pm$0.01 & -516.44$\pm$72.51 & 11.04$\pm$0.36 & 2.13$\pm$0.07 & 3342.38$\pm$104.14 s\\
 & ARTELT & - & - & - & - & - & - & - & - & - s\\
 & \our{} & \textbf{1.00$\pm$0.00} & \textbf{1.00$\pm$0.00} & \textbf{0.98$\pm$0.01} & 1.13$\pm$0.01 & 0.03$\pm$0.01 & \textbf{43.87$\pm$2.38} & 8.78$\pm$0.29 & \textbf{1.42$\pm$0.05} & 25.09$\pm$0.40 s\\
\midrule
\multirow{5}{*}{Wine} & CBCE & \textbf{1.00$\pm$0.00} & 0.96$\pm$0.04 & 0.37$\pm$0.05 & 1.11$\pm$0.03 & 0.03$\pm$0.02 & 1.71$\pm$1.93 & 4.05$\pm$0.22 & 1.31$\pm$0.06 & \textbf{0.01$\pm$0.00} s\\
 & CEGP & 0.98$\pm$0.02 & 0.42$\pm$0.05 & 0.06$\pm$0.08 & 1.12$\pm$0.03 & 0.04$\pm$0.01 & -1.36$\pm$3.40 & 1.31$\pm$0.18 & \textbf{0.48$\pm$0.07} & 486.70$\pm$46.60 s\\
 & CEM & \textbf{1.00$\pm$0.00} & 0.44$\pm$0.06 & 0.00$\pm$0.00 & 1.40$\pm$0.05 & -0.02$\pm$0.01 & -17.90$\pm$4.72 & \textbf{1.24$\pm$0.09} & 0.67$\pm$0.06 & 117.67$\pm$3.86 s\\
 & WACH & \textbf{1.00$\pm$0.00} & 0.80$\pm$0.16 & 0.01$\pm$0.01 & 1.29$\pm$0.10 & 0.00$\pm$0.02 & -10.69$\pm$7.38 & 2.07$\pm$0.23 & 0.80$\pm$0.06 & 224.63$\pm$6.95 s\\
 & ARTELT & - & - & - & - & - & - & - & - & - s\\
 & \our{} & \textbf{1.00$\pm$0.00} & \textbf{0.97$\pm$0.03} & \textbf{0.99$\pm$0.01} & 1.01$\pm$0.01 & 0.09$\pm$0.01 & \textbf{9.79$\pm$0.59} & 1.71$\pm$0.10 & 0.55$\pm$0.04 & 10.32$\pm$0.73 s\\
\bottomrule
\end{tabular}
\end{scriptsize}
\end{sc}
\end{center}
\end{table*}
\end{landscape}

\begin{landscape}
\begin{table*}[t]
\centering
\caption{Detailed Comparative Results of Probabilistically Plausible Counterfactual Explanation Methods for \textbf{Neural Oblivious Decision Ensembles} classifier. We offer a detailed comparison of our method with other established approaches using various datasets. This evaluation presents both the mean values and the standard deviations obtained through five-fold cross-validation.}
\label{tab:ours_vs_all_cv_node}

\begin{center}
\begin{sc}
\begin{scriptsize}
\begin{tabular}{l|l|rrrrrrrrr}
\toprule
Dataset & Method & Coverage $\uparrow$ & Validity $\uparrow$ & Prob. Plaus. $\uparrow$ & LOF & IsoForest & Log Dens. $\uparrow$ & L1 $\downarrow$ & L2 $\downarrow$ & Time $\downarrow$ \\
\midrule
\multirow{6}{*}{Moons} & CBCE & \textbf{1.00$\pm$0.00} & \textbf{1.00$\pm$0.00} & 0.50$\pm$0.01 & 1.04$\pm$0.03 & 0.02$\pm$0.01 & 0.90$\pm$1.11 & 0.62$\pm$0.06 & 0.48$\pm$0.04 & \textbf{0.41$\pm$0.09} s\\
& CEGP & - & - & - & - & - & - & - & - & - s\\
 & CEM & \textbf{0.99$\pm$0.01} & \textbf{1.00$\pm$0.00} & 0.07$\pm$0.02 & 2.00$\pm$0.09 & -0.07$\pm$0.01 & -4.36$\pm$2.79 & 0.61$\pm$0.03 & 0.57$\pm$0.02 & 1011.68$\pm$15.44 s\\
 & WACH & \textbf{1.00$\pm$0.00} & \textbf{1.00$\pm$0.00} & 0.02$\pm$0.02 & 1.29$\pm$0.04 & 0.01$\pm$0.00 & -0.48$\pm$0.33 & \textbf{0.33$\pm$0.01} & \textbf{0.26$\pm$0.01} & 1736.54$\pm$19.44 s\\
 & ARTELT & - & - & - & - & - & - & - & - & - s\\
 & \our{} & \textbf{1.00$\pm$0.00} & \textbf{1.00$\pm$0.00} & \textbf{0.99$\pm$0.00} & 0.99$\pm$0.01 & 0.03$\pm$0.01 & \textbf{1.60$\pm$0.03} & 0.42$\pm$0.03 & 0.33$\pm$0.03 & 39.68$\pm$4.48 s\\
\midrule
\multirow{6}{*}{Law} & CBCE & \textbf{1.00$\pm$0.00} & \textbf{1.00$\pm$0.00} & 0.48$\pm$0.36 & 1.05$\pm$0.02 & 0.04$\pm$0.02 & 1.27$\pm$0.43 & 0.61$\pm$0.03 & 0.40$\pm$0.02 & \textbf{1.08$\pm$0.04} s\\
& CEGP & - & - & - & - & - & - & - & - & - s\\
 & CEM & 0.97$\pm$0.01 & \textbf{1.00$\pm$0.00} & 0.29$\pm$0.01 & 1.27$\pm$0.02 & -0.02$\pm$0.00 & -0.41$\pm$0.14 & \textbf{0.33$\pm$0.02} & 0.31$\pm$0.01 & 2103.30$\pm$31.69 s\\
 & WACH & \textbf{0.99$\pm$0.01} & \textbf{1.00$\pm$0.00} & 0.45$\pm$0.06 & 1.08$\pm$0.01 & 0.03$\pm$0.00 & 1.16$\pm$0.06 & 0.41$\pm$0.01 & 0.29$\pm$0.00 & 4401.60$\pm$240.03 s\\
 & ARTELT & - & - & - & - & - & - & - & - & - s\\
 & \our{} & \textbf{1.00$\pm$0.00} & \textbf{1.00$\pm$0.00} & \textbf{1.00$\pm$0.00} & 1.03$\pm$0.00 & 0.07$\pm$0.00 & \textbf{2.04$\pm$0.01} & 0.38$\pm$0.01 & \textbf{0.23$\pm$0.01} & 61.67$\pm$2.88 s\\
\midrule
\multirow{6}{*}{Audit} & CBCE & \textbf{1.00$\pm$0.00} & \textbf{1.00$\pm$0.00} & 0.79$\pm$0.28 & 11.80$\pm$20.40 & 0.14$\pm$0.00 & 54.92$\pm$3.92 & 2.55$\pm$0.19 & 1.24$\pm$0.10 & \textbf{0.38$\pm$0.08} s\\
& CEGP & - & - & - & - & - & - & - & - & - s\\
 & CEM & 0.52$\pm$0.03 & \textbf{1.00$\pm$0.00} & 0.00$\pm$0.01 & 1.52$\cdot\text{10}^8\pm$1.43$\cdot\text{10}^8$ & 0.11$\pm$0.00 & 12.63$\pm$14.72 & \textbf{1.06$\pm$0.10} & \textbf{0.56$\pm$0.04} & 906.63$\pm$24.07 s\\
 & WACH & 0.98$\pm$0.03 & \textbf{1.00$\pm$0.00} & 0.03$\pm$0.04 & 6.90$\cdot\text{10}^7\pm$7.94$\cdot\text{10}^7$ & 0.06$\pm$0.01 & -33.50$\pm$50.95 & 1.59$\pm$0.14 & 0.97$\pm$0.04 & 2347.91$\pm$227.09 s\\
 & ARTELT & - & - & - & - & - & - & - & - & - s\\
 & \our{} & \textbf{1.00$\pm$0.00} & \textbf{1.00$\pm$0.00} & \textbf{0.99$\pm$0.01} & 2.32$\cdot\text{10}^7\pm$2.99$\cdot\text{10}^7$ & 0.09$\pm$0.01 & \textbf{51.67$\pm$4.53} & 2.06$\pm$0.12 & 0.80$\pm$0.09 & 80.34$\pm$21.47 s\\
\midrule
\multirow{6}{*}{Heloc} & CBCE & \textbf{1.00$\pm$0.00} & \textbf{1.00$\pm$0.00} & 0.55$\pm$0.03 & 1.09$\pm$0.08 & 0.08$\pm$0.03 & 28.88$\pm$4.06 & 2.85$\pm$0.40 & 0.82$\pm$0.11 & \textbf{17.53$\pm$0.92} s\\
& CEGP & - & - & - & - & - & - & - & - & - s\\
 & CEM & 0.94$\pm$0.01 & \textbf{1.00$\pm$0.00} & 0.10$\pm$0.01 & 1.35$\pm$0.01 & 0.05$\pm$0.01 & 9.00$\pm$3.61 & \textbf{0.47$\pm$0.08} & \textbf{0.29$\pm$0.02} & 14772.66$\pm$226.75 s\\
 & WACH & 0.96$\pm$0.03 & \textbf{1.00$\pm$0.00} & 0.10$\pm$0.02 & 2.12$\cdot\text{10}^8\pm$4.23$\cdot\text{10}^8$ & 0.05$\pm$0.01 & 10.75$\pm$10.46 & 0.85$\pm$0.05 & 0.36$\pm$0.04 & 37254.33$\pm$3666.87 s\\
 & ARTELT & - & - & - & - & - & - & - & - & - s\\
 & \our{} & \textbf{1.00$\pm$0.00} & 0.94$\pm$0.01 & \textbf{1.00$\pm$0.00} & 1.08$\pm$0.00 & 0.09$\pm$0.00 & \textbf{31.85$\pm$0.41} & 1.02$\pm$0.06 & 0.28$\pm$0.02 & 126.05$\pm$33.10 s\\
\midrule
\multirow{6}{*}{Blobs} & CBCE & \textbf{1.00$\pm$0.00} & \textbf{1.00$\pm$0.00} & 0.27$\pm$0.15 & 1.02$\pm$0.03 & 0.03$\pm$0.02 & -35.52$\pm$15.68 & 0.95$\pm$0.01 & 0.72$\pm$0.01 & \textbf{0.77$\pm$0.08} s\\
& CEGP & - & - & - & - & - & - & - & - & - s\\
 & CEM & 0.90$\pm$0.04 & \textbf{1.00$\pm$0.00} & 0.00$\pm$0.00 & 3.10$\pm$0.09 & -0.11$\pm$0.00 & -24.98$\pm$10.42 & 0.58$\pm$0.02 & 0.52$\pm$0.02 & 1329.91$\pm$14.61 s\\
 & WACH & 0.94$\pm$0.02 & \textbf{1.00$\pm$0.00} & 0.00$\pm$0.00 & 2.65$\pm$0.11 & -0.08$\pm$0.00 & -11.76$\pm$3.58 & \textbf{0.37$\pm$0.02} & \textbf{0.33$\pm$0.01} & 2406.40$\pm$59.59 s\\
 & ARTELT & - & - & - & - & - & - & - & - & - s\\
 & \our{} & \textbf{1.00$\pm$0.00} & \textbf{1.00$\pm$0.00} & \textbf{1.00$\pm$0.00} & 1.02$\pm$0.02 & 0.04$\pm$0.00 & \textbf{2.94$\pm$0.04} & 0.65$\pm$0.01 & 0.47$\pm$0.01 & 47.69$\pm$5.21 s\\
\midrule
\multirow{6}{*}{Digits} & CBCE & \textbf{1.00$\pm$0.00} & \textbf{1.00$\pm$0.00} & 0.18$\pm$0.04 & 1.02$\pm$0.06 & 0.04$\pm$0.00 & 24.00$\pm$4.14 & 16.27$\pm$0.65 & 3.09$\pm$0.03 & \textbf{3.12$\pm$0.24} s\\
& CEGP & - & - & - & - & - & - & - & - & - s\\
 & CEM & \textbf{1.00$\pm$0.00} & \textbf{1.00$\pm$0.00} & 0.03$\pm$0.01 & 1.32$\pm$0.01 & -0.02$\pm$0.00 & -39.45$\pm$10.78 & 4.07$\pm$0.14 & 1.44$\pm$0.03 & 5451.83$\pm$19.15 s\\
 & WACH & \textbf{1.00$\pm$0.00} & \textbf{1.00$\pm$0.00} & 0.16$\pm$0.03 & 1.12$\pm$0.01 & 0.02$\pm$0.01 & 7.02$\pm$5.00 & \textbf{2.93$\pm$0.05} & \textbf{1.13$\pm$0.01} & 15376.44$\pm$290.44 s\\
 & ARTELT & - & - & - & - & - & - & - & - & - s\\
 & \our{} & \textbf{1.00$\pm$0.00} & \textbf{1.00$\pm$0.00} & \textbf{1.00$\pm$0.00} & 1.15$\pm$0.01 & 0.02$\pm$0.01 & \textbf{43.97$\pm$1.95} & 7.76$\pm$0.20 & 1.36$\pm$0.03 & 69.45$\pm$3.94 s\\
\midrule
\multirow{6}{*}{Wine} & CBCE & \textbf{1.00$\pm$0.00} & \textbf{1.00$\pm$0.00} & 0.39$\pm$0.03 & 1.11$\pm$0.04 & 0.02$\pm$0.02 & 2.04$\pm$1.86 & 4.05$\pm$0.22 & 1.31$\pm$0.06 & \textbf{0.10$\pm$0.07} s\\
& CEGP & - & - & - & - & - & - & - & - & - s\\
 & CEM & \textbf{1.00$\pm$0.00} & \textbf{1.00$\pm$0.00} & 0.00$\pm$0.00 & 1.34$\pm$0.04 & 0.00$\pm$0.01 & -21.02$\pm$14.00 & 1.11$\pm$0.19 & 0.65$\pm$0.06 & 225.63$\pm$11.18 s\\
 & WACH & \textbf{0.99$\pm$0.01} & \textbf{1.00$\pm$0.00} & 0.08$\pm$0.06 & 1.12$\pm$0.04 & 0.04$\pm$0.01 & -1.05$\pm$2.49 & \textbf{1.10$\pm$0.04} & 0.59$\pm$0.02 & 459.91$\pm$13.22 s\\
 & ARTELT & - & - & - & - & - & - & - & - & - s\\
 & \our{} & \textbf{1.00$\pm$0.00} & \textbf{1.00$\pm$0.00} & \textbf{1.00$\pm$0.00} & 1.01$\pm$0.01 & 0.09$\pm$0.00 & \textbf{9.82$\pm$0.66} & 1.69$\pm$0.09 & \textbf{0.55$\pm$0.03} & 19.39$\pm$1.06 s\\
\bottomrule
\end{tabular}
\end{scriptsize}
\end{sc}
\end{center}
\end{table*}
\end{landscape}

\begin{table}[t]
\centering
\caption{Detailed Ablation Study on Loss Function Selection. This table includes additional metrics: LOF and IsoForest}
%  We compare the impact of using our proposed loss function versus Binary Cross Entropy (BCE) on counterfactual explanation metrics across various datasets. The results demonstrate that our loss function is more effective in producing closer and more plausible counterfactuals.
\label{tab:ablation_loss_cv}

\begin{center}
\begin{sc}
\begin{scriptsize}
\begin{tabular}{l|l|rrrrrrrr}
\toprule
Dataset & Loss & Cov.$\uparrow$ & Val.$\uparrow$ & Prob. Plaus.$\uparrow$ & LOF & IsoForest & Log Dens.$\uparrow$ & L1$\downarrow$ & L2$\downarrow$ \\
\midrule
\multirow{2}{*}{Moons} & Ours & \textbf{1.00$\pm$0.00} & \textbf{1.00$\pm$0.00} & \textbf{1.00$\pm$0.00} & 1.01$\pm$0.02 & 0.04$\pm$0.01 & 1.69$\pm$0.07 & \textbf{0.45$\pm$0.01} & \textbf{0.36$\pm$0.01} \\
 & BCE & \textbf{1.00$\pm$0.00} & \textbf{1.00$\pm$0.00} & \textbf{0.99$\pm$0.00} & 1.04$\pm$0.03 & -0.01$\pm$0.01 & \textbf{1.74$\pm$0.09} & 0.89$\pm$0.03 & 0.69$\pm$0.02 \\
\midrule
\multirow{2}{*}{Law} & Ours & \textbf{1.00$\pm$0.00} & \textbf{1.00$\pm$0.00} & \textbf{1.00$\pm$0.00} & 1.03$\pm$0.00 & 0.07$\pm$0.00 & \textbf{2.05$\pm$0.02} & \textbf{0.37$\pm$0.01} & \textbf{0.23$\pm$0.01} \\
 & BCE & \textbf{1.00$\pm$0.00} & \textbf{1.00$\pm$0.00} & \textbf{0.98$\pm$0.01} & 0.99$\pm$0.00 & 0.01$\pm$0.01 & 1.67$\pm$0.01 & 0.97$\pm$0.02 & 0.60$\pm$0.01 \\
\midrule
\multirow{2}{*}{Audit} & Ours & \textbf{1.00$\pm$0.00} & \textbf{0.99$\pm$0.01} & \textbf{0.99$\pm$0.01} & 4.25$\cdot\text{10}^7\pm$9.32$\cdot\text{10}^7$ & 0.08$\pm$0.01 & 51.64$\pm$4.53 & \textbf{2.04$\pm$0.15} & \textbf{0.79$\pm$0.12} \\
 & BCE & \textbf{1.00$\pm$0.00} & \textbf{0.99$\pm$0.01} & \textbf{0.98$\pm$0.01} & 3.59$\cdot\text{10}^8\pm$1.16$\cdot\text{10}^8$ & 0.09$\pm$0.01 & \textbf{52.54$\pm$4.54} & 3.01$\pm$0.20 & 1.25$\pm$0.10 \\
\midrule
\multirow{2}{*}{Heloc} & Ours & \textbf{1.00$\pm$0.00} & \textbf{1.00$\pm$0.00} & \textbf{1.00$\pm$0.00} & 6.47$\cdot\text{10}^7\pm$2.16$\cdot\text{10}^7$ & 0.07$\pm$0.00 & \textbf{32.42$\pm$0.34} & \textbf{0.90$\pm$0.03} & \textbf{0.23$\pm$0.01} \\
 & BCE & \textbf{1.00$\pm$0.00} & \textbf{1.00$\pm$0.00} & \textbf{0.99$\pm$0.00} & 1.67$\cdot\text{10}^8\pm$6.18$\cdot\text{10}^7$ & 0.05$\pm$0.01 & 32.11$\pm$0.45 & 2.77$\pm$0.13 & 0.78$\pm$0.05 \\
\midrule
\multirow{2}{*}{Blobs} & Ours & \textbf{1.00$\pm$0.00} & \textbf{1.00$\pm$0.00} & \textbf{1.00$\pm$0.00} & 1.01$\pm$0.01 & 0.04$\pm$0.01 & \textbf{3.00$\pm$0.11} & \textbf{0.69$\pm$0.05} & \textbf{0.50$\pm$0.04} \\
 & CE & \textbf{1.00$\pm$0.00} & \textbf{1.00$\pm$0.00} & 0.93$\pm$0.01 & 1.05$\pm$0.02 & 0.03$\pm$0.01 & 2.85$\pm$0.04 & 0.82$\pm$0.02 & 0.60$\pm$0.01 \\
\midrule
\multirow{2}{*}{Digits} & Ours & \textbf{1.00$\pm$0.00} & \textbf{1.00$\pm$0.00} & \textbf{1.00$\pm$0.00} & 1.12$\pm$0.01 & 0.03$\pm$0.01 & \textbf{44.42$\pm$1.87} & \textbf{8.27$\pm$0.24} & \textbf{1.33$\pm$0.04} \\
 & CE & \textbf{1.00$\pm$0.00} & \textbf{1.00$\pm$0.00} & \textbf{1.00$\pm$0.00} & 1.12$\pm$0.01 & 0.02$\pm$0.01 & 44.18$\pm$2.09 & 12.67$\pm$0.15 & 2.13$\pm$0.04 \\
\midrule
\multirow{2}{*}{Wine} & Ours & \textbf{1.00$\pm$0.00} & \textbf{1.00$\pm$0.00} & \textbf{1.00$\pm$0.00} & 1.01$\pm$0.01 & 0.09$\pm$0.01 & \textbf{9.72$\pm$0.62} & \textbf{1.65$\pm$0.09} & \textbf{0.53$\pm$0.04} \\
 & CE & \textbf{1.00$\pm$0.00} & \textbf{1.00$\pm$0.00} & \textbf{0.99$\pm$0.01} & 1.06$\pm$0.03 & 0.02$\pm$0.01 & 9.29$\pm$0.71 & 3.87$\pm$0.18 & 1.29$\pm$0.06 \\
\bottomrule
\end{tabular}
\end{scriptsize}
\end{sc}
\end{center}
\end{table}

% \subsection{\our{} vs. Artelt on other datasets}

\section{Datasets}
\label{apx:datasets}
In Tab. \ref{tab:dataset_description}, we provide detailed descriptions of the datasets utilized in our study: Moons, Law\footnote{\url{https://www.kaggle.com/datasets/danofer/law-school-admissions-bar-passage}}, Audit\footnote{\url{https://archive.ics.uci.edu/dataset/475/audit+data}}, Heloc\footnote{\url{https://community.fico.com/s/explainable-machine-learning-challenge}}, Wine\footnote{\url{https://archive.ics.uci.edu/dataset/109/wine}}, Blobs and Digits\footnote{\url{https://archive.ics.uci.edu/dataset/80/optical+recognition+of+handwritten+digits}}. The Moons dataset is an artificially generated set comprising two interleaving half-circles. It includes a standard deviation of Gaussian noise set at 0.01. The Law dataset originates from the Law School Admissions Council (LSAC) and is referred to in the literature as the Law School Admissions dataset \cite{Wightman98}. For our analysis, we selected three features that exhibit the highest correlation with the target variable: entrance exam scores (LSAT), grade-point average (GPA), and first-year average grade (FYA). The Audit dataset, which encompasses comprehensive one-year non-confidential data of firms in the years 2015 to 2016, is collected from the Auditor Office of India to build a predictor for classifying suspicious firms. The Heloc dataset, initially utilized in the 'FICO xML Challenge', consists of Home Equity Line of Credit (HELOC) applications submitted by real homeowners. This dataset comprises various numeric features that encapsulate information from the applicant's credit report. The primary objective is to predict whether the applicant will repay their HELOC account within a two-year period. This prediction is instrumental in determining the applicant's qualification for a line of credit. The Wine dataset comprises chemical analysis results for wines originating from the same region in Italy, produced from three distinct cultivars. This analysis quantified 13 different constituents present in each of the three wine varieties. The Blobs dataset is an artificially generated isotropic Gaussian blobs, characterized by equal variance. The Digits dataset is utilized for the optical recognition of handwritten digits. It consists of 32x32 bitmap images that are segmented into non-overlapping 4x4 blocks. Within each block, the count of 'on' pixels is recorded, resulting in an 8x8 input matrix. Each element of this matrix is an integer between 0 and 16.

\begin{table}[t]
\centering
\caption{Dataset Characteristics and Model Performances. This table provides an overview of the datasets used in our experiments, including the number of samples ($N$), number of features ($d$), number of classes ($C$), accuracy of Logistic Regression (LR Acc.), Multi-Layer Perceptron (MLP Acc.), and the log density of the Masked Autoregressive Flow (MAF Log Dens.).}
\label{tab:dataset_description}

\begin{center}
\begin{sc}
\begin{tabular}{l|rrrrrr}
\toprule
Dataset & $N$ & $d$ & $C$ & LR Acc. & MLP Acc. & MAF Log Dens.\\
\midrule
Moons &  1,024 & 2  & 2 & 0.85 & 0.98 & 1.38 \\
Law   &  2,220 & 3  & 2 & 0.75 & 0.75 &  1.23 \\
Audit &    610 & 23 & 2 & 0.95 & 0.98 &  48.15 \\
Heloc & 10,459 & 23 & 2 & 0.70 & 0.70 &  28.67 \\
\midrule
Wine  & 178  & 13   & 3 & 0.90 & 0.97 & 7.21 \\
Blobs & 1,500 &  2   & 3 & 1.00 & 1.00 & 2.58 \\
Digits& 5,620 & 64   & 10& 0.94 & 0.95 & 35.80 \\
\bottomrule
\end{tabular}
\end{sc}
\end{center}
\end{table}

\section{Density Estimator - Additional Results}
\label{sec:density_estimator}
This experiment assesses the efficacy of Normalizing Flow models, particularly in high-dimensional datasets, against traditional Kernel Density Estimation (KDE). We compared KDE with three Normalizing Flow architectures: RealNVP, NICE, and Masked Autoregressive Flow (MAF). The mean log density results for the test datasets are detailed in Tab. \ref{tab:density_estimator}. For lower-dimensional datasets like Moons, Law and Blobs, KDE and MAF show comparable performance, significantly outperforming RealNVP and NICE. However, in high-dimensional datasets such as Audit, Heloc and Digits, MAF demonstrates a substantial advantage over the other density estimators. These findings reinforce our proposition to employ Normalizing Flows, especially the MAF architecture, as effective density estimators in our method.

\begin{table}[ht]
\centering
\caption{Comparative Analysis of Density Estimators. We present the mean log density results for KDE, RealNVP, NICE, and MAF across various datasets. It highlights the superior performance of MAF in high-dimensional datasets (Audit and Heloc) and its comparable efficacy to KDE in lower-dimensional datasets (Moons and Law), underscoring the effectiveness of MAF as a density estimator in our method.}
\label{tab:density_estimator}

\begin{center}
\begin{sc}
\begin{tabular}{l|rrrr}
\toprule
Dataset & KDE & RealNVP & NICE & MAF \\
\midrule
Moons & 0.95$\pm$0.01 & -1.85$\pm$0.00 & -1.86$\pm$0.00 & \textbf{1.38$\pm$0.07} \\
Law & 1.16$\pm$0.03 & -2.79$\pm$0.00 & -2.80$\pm$0.00 & \textbf{1.23$\pm$0.05} \\
Audit & 10.75$\pm$28.15 & -21.25$\pm$0.13 & -21.34$\pm$0.22 & \textbf{48.15$\pm$8.41} \\
Heloc & 22.44$\pm$0.32 & -21.12$\pm$0.00 & -21.19$\pm$0.00 & \textbf{28.67$\pm$0.42} \\
Wine & 5.95$\pm$0.57 & -12.05$\pm$0.02 & -12.08$\pm$0.01 & \textbf{7.21$\pm$0.80} \\
Blobs & 2.10$\pm$0.02 & -1.84$\pm$0.00 & -1.84$\pm$0.00 & \textbf{2.58$\pm$0.05} \\
Digits & 22.66$\pm$1.14 & -59.33$\pm$0.05 & -59.79$\pm$0.09 & \textbf{35.80$\pm$3.30} \\
\bottomrule
\end{tabular}
\end{sc}
\end{center}
\end{table}

\section{Implementation details}
In the implementation of our experiments, we utilized Python~\citep{van1995python} as the primary programming language. The core of our computational framework was PyTorch~\citep{pytorch}, a popular open-source machine learning library. A key feature of our implementation was the gradient optimization approach, designed to be executed in batches. This approach was particularly effective, allowing us to process entire test sets in a single batch. Our experiments were conducted on an M1 Apple Silicon CPU paired with 16GB of RAM. This hardware setup was more than sufficient for our experiment needs, providing enough capacity for the computational demands of our algorithm while ensuring fast processing speeds.

\end{document}